\documentclass[10pt,twocolumn,letterpaper]{article}
\usepackage[pagenumbers]{cvpr}
\usepackage{microtype}
\usepackage{cuted}
\usepackage{placeins}
\usepackage{mathtools}
\usepackage{xspace}
\usepackage{url}
\usepackage{colortbl}
\usepackage{tikz}

\newcommand{\benchmark}{Code2Video-Compass\xspace}
\newcommand{\bpstart}[1]{\vspace{2pt}\noindent\textbf{#1.}}
\newcommand{\metricstart}[1]{\par\addvspace{1pt}\noindent\hspace*{\parindent}\mbox{\textbf{\textit{#1.}}}\hspace{0.5em}\ignorespaces}
\definecolor{closedsourcecolor}{HTML}{4C5665}
\definecolor{openweightcolor}{HTML}{3C735B}
\newcommand{\modelentry}[2]{\makebox[11pt][l]{\raisebox{-1.2pt}{#1}}#2}
\newcommand{\modelLogoOpenAI}{\includegraphics[width=7.5pt,height=7.5pt,keepaspectratio]{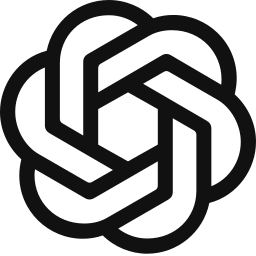}}
\newcommand{\modelLogoClaude}{\includegraphics[width=7.5pt,height=7.5pt,keepaspectratio]{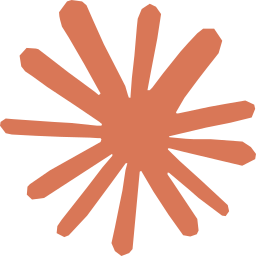}}
\newcommand{\modelLogoGemini}{\includegraphics[width=7.5pt,height=7.5pt,keepaspectratio]{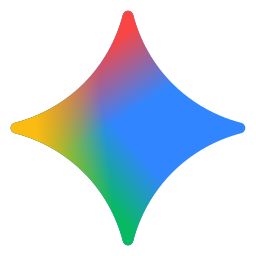}}
\newcommand{\modelLogoQwen}{\includegraphics[width=7.5pt,height=7.5pt,keepaspectratio]{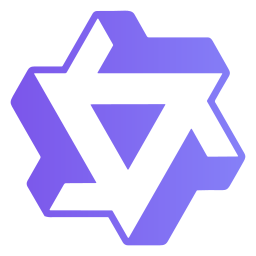}}
\newcommand{\modelLogoKimi}{\includegraphics[width=7.5pt,height=7.5pt,keepaspectratio]{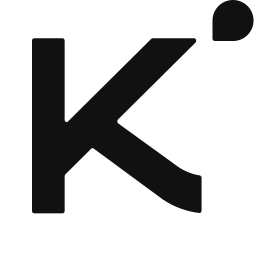}}
\newcommand{\modelLogoGLM}{\includegraphics[width=7.5pt,height=7.5pt,keepaspectratio]{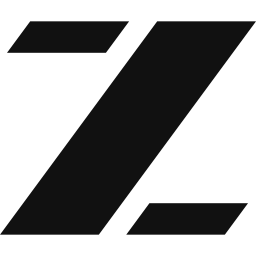}}
\newcommand{\modelLogoDeepSeek}{\includegraphics[width=7.5pt,height=7.5pt,keepaspectratio]{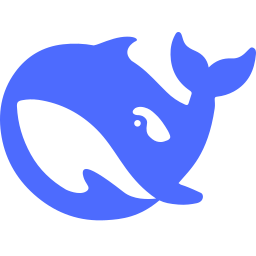}}
\newcommand{\modelLogoMiniMax}{\includegraphics[width=7.5pt,height=7.5pt,keepaspectratio]{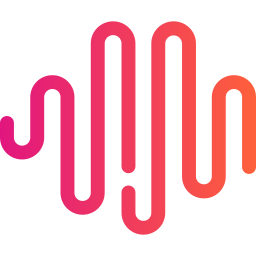}}
\newcommand{\modelLogoMiMo}{\includegraphics[width=7.5pt,height=7.5pt,keepaspectratio]{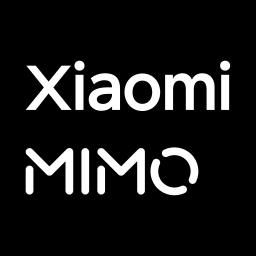}}
\newcommand{\modelgroup}[2]{\textcolor{#1}{\textbf{\textit{#2}}}}

\DeclareRobustCommand{\frameinput}{\raisebox{-2pt}{\includegraphics[width=8pt,height=8pt]{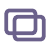}}}

\definecolor{HdrOperational}{HTML}{35840D}
\definecolor{HdrVideo}{HTML}{35840D}
\definecolor{HdrAlignment}{HTML}{35840D}
\definecolor{HdrImageFidelity}{HTML}{35840D}
\definecolor{HdrClipFidelity}{HTML}{35840D}

\definecolor{HdrSubjective}{HTML}{ECA138}
\definecolor{HdrFollowing}{HTML}{ECA138}
\definecolor{HdrOverall}{HTML}{7064A8}
\definecolor{BestBg}{HTML}{DED8EE}
\definecolor{SecondBg}{HTML}{F1EDF8}
\definecolor{OpenWeightBg}{HTML}{F1F6F3}
\definecolor{ClosedSourceBg}{HTML}{F8F4EC}
\definecolor{AvgRankTint}{HTML}{B6BCE3}
\newcommand{\avgrankcell}[2]{\cellcolor{AvgRankTint!#1}#2}
\newcommand{\bestscore}[1]{\textbf{#1}}
\newcommand{\secondscore}[1]{\underline{#1}}

\definecolor{cvprblue}{rgb}{0.21,0.49,0.74}
\usepackage[pagebackref,breaklinks,colorlinks,allcolors=cvprblue]{hyperref}
\hypersetup{
  pdftitle={Code2Video-Compass: A Comprehensive Benchmark for General-Purpose Code-Based Video Generation across T2V, I2V, and V2V},
  pdfauthor={Zhenghan Chen, Zekai Shao, Xin Lin, Leting Cheng, and Siming Chen}
}
\title{Code2Video-Compass: A Comprehensive Benchmark for General-Purpose Code-Based Video Generation across T2V, I2V, and V2V}
\author{\begin{tabular}{c}
    Zhenghan Chen\textsuperscript{1},
    Zekai Shao\textsuperscript{1,$\dagger$},
    Xin Lin\textsuperscript{2},
    Leting Cheng\textsuperscript{1}, and
    Siming Chen\textsuperscript{1,*}\\
    \textsuperscript{1}Fudan University\qquad
    \textsuperscript{2}Sun Yat-sen University\\
    {\small \textsuperscript{$\dagger$}Project Lead,\quad
    \textsuperscript{*}Corresponding author}\\
    {\large Project Repository:
      \href{https://code2video-compass.github.io/}{https://code2video-compass.github.io/}}
  \end{tabular}}
\begin{document}
\raggedbottom
\maketitle
\begin{strip}
  \centering
  \vspace{-40pt}
  \includegraphics[width=\textwidth]{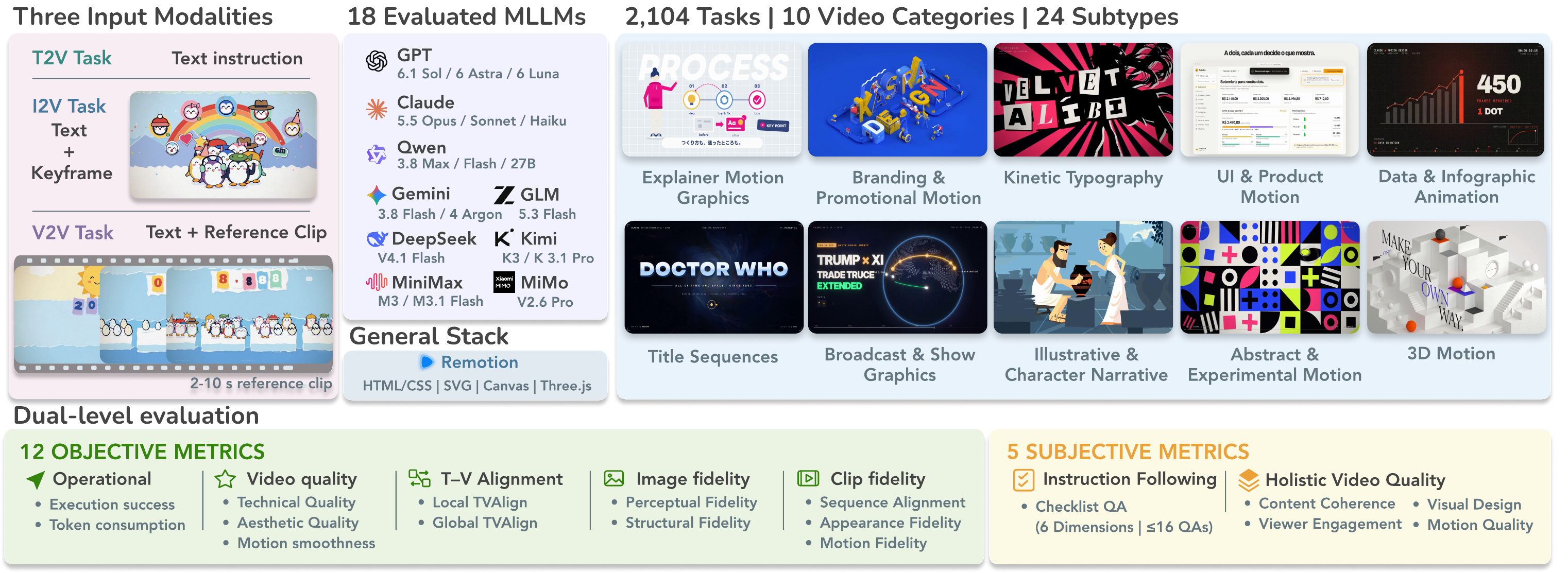}
  \captionsetup{hypcap=false}
  \captionof{figure}{\textbf{Overview of Code2Video-Compass.}
  The benchmark evaluates 18 multimodal large language models (MLLMs) on code-based text-to-video (T2V), image-to-video (I2V), and video-to-video (V2V) generation.
  Models generate executable code from text and optional visual references using a general-purpose stack supporting HTML/CSS, SVG, Canvas, and Three.js, with Remotion rendering the resulting videos.
  It comprises 2,104 tasks spanning 10 categories and 24 subtypes, evaluated through a comprehensive dual-level framework combining 12 objective metrics with five subjective metrics.}
  \label{fig:teaser}
\end{strip}

\begin{abstract}
Code-based video generation offers fine-grained control, targeted editing, and component reuse compared with direct video generation. This approach requires multimodal large language models (MLLMs) to translate visual and temporal requirements into executable programs that produce complete videos.
However, existing code-based benchmarks primarily target specialized tasks, leaving complete video creation across diverse creative categories and input modalities insufficiently evaluated.
We introduce \benchmark, a general-purpose benchmark for code-based video generation across text-to-video (T2V), image-to-video (I2V), and video-to-video (V2V) settings.
\benchmark comprises 2,104 tasks spanning 10 categories, 24 subtypes, and three complexity dimensions covering construction, dynamic animation, and state dependency.
It combines tasks curated from textual instructions and visual references with corresponding T2V, I2V, and V2V tasks derived from complete professional and user-created videos.
We propose a dual-level evaluation framework combining 12 objective metrics and five subjective metrics. Objective evaluation covers operational performance, video quality, text--video alignment, and reference fidelity, while subjective evaluation uses an MLLM judge to evaluate instruction following and holistic video quality.
Extensive evaluation of 18 recent proprietary and open-weight MLLMs from nine families reveals that even leading models face challenges in reliably generating high-quality complete videos aligned with textual and visual requirements, with performance generally declining as construction, dynamic animation, and state-dependency complexity increases.
We anticipate that \benchmark will advance reliable, coherent, and controllable code-based video generation and inspire broader progress in programmatic visual creation.
\end{abstract}

\section{Introduction}
\label{sec:introduction}
Recent advances in multimodal large language models (MLLMs)~\cite{gpt6astra,claudeopus55,gemini38flash,qwen38max,kimik3} are making it increasingly feasible to create complete videos through executable code, a task requiring world knowledge, spatial reasoning, creative judgment, and complex programming.
Compared with direct video generation, code-based approaches explicitly represent visual structures and animation parameters, enabling fine-grained control, targeted editing, and component reuse~\cite{designascode,lottiegpt,remotion}.
Generating complete videos that follow textual instructions requires models to plan coherent narratives and visuals, reliably generate lengthy code across interdependent files, and coordinate animations and evolving states across scenes and timelines.
Moreover, visual references introduce additional challenges~\cite{trip,10655955}, requiring models to reconstruct the referenced appearance or motion and integrate it into the complete video.

Existing benchmarks for direct video generation, including VBench~\cite{vbench}, EvalCrafter~\cite{evalcrafter}, and T2V-CompBench~\cite{t2vcompbench}, evaluate visual quality, semantic alignment, and compositional requirements, while LongAV-Compass~\cite{longavcompass} extends evaluation to longer sequences and multiple input conditions. Code-based tasks target specialized applications or representations, including Manim educational videos~\cite{code2video}, data narratives~\cite{datareel}, SVG animations~\cite{internsvg}, and web animation reconstruction~\cite{animation2code}. Similarly, HeyGen's Code2Video Benchmark~\cite{heygen} focuses on short motion graphics clips representing individual beats of product launch videos.
Nevertheless, as shown in Tab.~\ref{tab:benchmark_comparison}, existing code-based benchmarks remain limited in task coverage and input modalities, and lack comprehensive evaluation combining objective and subjective metrics.

To address these gaps, we introduce \textbf{\benchmark} (Fig.~\ref{fig:teaser}), a general-purpose benchmark for code-based text-to-video (T2V), image-to-video (I2V), and video-to-video (V2V) generation, where models create complete videos following textual instructions and, when provided, reconstruct and integrate visual references at appropriate temporal positions.
Guided by a taxonomy of motion-design practices, we combine independently constructed tasks based on curated textual instructions and visual references with tasks derived from complete professional and user-created videos.
This yields 2,104 tasks across 10 categories and 24 subtypes, each annotated along three complexity dimensions: construction, dynamic animation, and state dependency.
We further propose a comprehensive dual-level evaluation framework combining 12 objective metrics and five subjective metrics.
Objective evaluation measures execution success, token consumption, technical and aesthetic quality, motion smoothness, text--video alignment, and reference fidelity.
Subjective evaluation employs an MLLM-as-a-judge protocol to evaluate fine-grained instruction following through explicit checklists and evaluate holistic video quality across content coherence, visual design, motion quality, and viewer engagement.

In summary, our main contributions are as follows:
\begin{itemize}
    \item \textbf{General-Purpose Code-Based Video Generation.}
    We formulate unified T2V, I2V, and V2V tasks that require MLLMs to generate executable code for video generation.
    \item \textbf{Taxonomy-Driven Benchmark Dataset.}
    We curate 2,104 tasks spanning 10 categories and 24 subtypes, annotated along three complexity dimensions, each with three levels.
    \item \textbf{Comprehensive Dual-Level Evaluation Framework.}
    We integrate 12 objective metrics and five subjective metrics to evaluate code-based video generation capabilities.
    \item \textbf{Extensive MLLM Benchmarking and Insights.} We evaluate 18 proprietary and open-weight MLLMs, revealing that video quality and execution reliability do not always align, with performance declining as complexity increases.
\end{itemize}

\begin{table*}[t]
  \centering
  \captionsetup{skip=7pt}

\caption{\textbf{Comparison of representative video generation benchmarks.}
Source denotes Real (derived from existing works), Synthetic (constructed without such works), or Hybrid (both).
T, I, and V denote text, image, and video inputs; Duration is the average video length in minutes.
Dual combines objective and subjective evaluation.
Avg. Tokens uses the Qwen3 tokenizer~\cite{qwen3}, while event counts follow T2AV-Compass~\cite{t2avcompass}.
$^{\dagger}$ indicates statistics from partial datasets; N/A and -- denote visual-only input and unavailable data, respectively.
}

  \label{tab:benchmark_comparison}
  \begingroup
  \footnotesize
  \setlength{\tabcolsep}{2pt}
  \renewcommand{\arraystretch}{1.20}
  \begin{tabular*}{\textwidth}{@{\extracolsep{\fill}}*{11}{c}@{}}
    \toprule
    \textbf{Benchmark} & \textbf{\#Items} & \textbf{Source} & \textbf{Approach} &
    \textbf{Task Scope} & \textbf{Input} & \textbf{Output} &
    \textbf{Duration} &
    \textbf{Avg. Tokens / Events} &
    \textbf{Evaluation} & \textbf{\#Metrics} \\
    \midrule
    \multicolumn{11}{@{}l}{\textit{Direct Video Generation}} \\
    VBench~\cite{vbench} & 946 & Synthetic & Direct & General & T & Video &
    $<1$ & 10 / 1.06 & Objective & 16 \\
    VABench~\cite{vabench} & 1,299 & Synthetic & Direct & General & T / I & Video &
    $<1$ & 50 / 2.31 & Dual & 15 \\
    T2AV-Compass~\cite{t2avcompass} & 500 & Hybrid & Direct & General & T & Video &
    $<1$ & 154 / 3.61 & Dual & 13 \\
    LongAV-Compass~\cite{longavcompass} & 284 & Hybrid & Direct & General & T / I / V & Video &
    $>1$ & 209 / 6.76 & Dual & 16 \\
    \midrule
    \multicolumn{11}{@{}l}{\textit{Code-Based Generation}} \\
    MMMC (Code2Video)~\cite{code2video} & 456 & Real & Manim & Education & T & Video &
    $>1$ & 8$^{\dagger}$ / -- & Dual & 8 \\
    SArena-Animation~\cite{internsvg} & 1,008 & Hybrid & SVG & Animation & T / V & SVG &
    $<1$ & 65 / -- & Objective & 8 \\
    Animation2Code~\cite{animation2code} & 1,069 & Real & Web & Animation & V & Video &
    $<1$ & N/A / -- & Objective & 3 \\
    HeyGen Code2Video~\cite{heygen} & 168 & Real & General & Product & T & Video &
    $<1$ & 471$^{\dagger}$ / -- & Subjective & 5 \\
    \midrule
    \textbf{Code2Video-Compass (Ours)} & \textbf{2,104} & \textbf{Hybrid} & \textbf{General} &
    \textbf{General} & \textbf{T / I / V} & \textbf{Video} & \textbf{\boldmath$>1$} & \textbf{594 / 9.46} &
    \textbf{Dual} & \textbf{17} \\
    \bottomrule
  \end{tabular*}
  \endgroup
\end{table*}

\section{Related Work}
\label{sec:related}

\subsection{Direct Video Generation}

Benchmarks for direct video generation have established diverse evaluation tasks and frameworks for assessing model capabilities. VBench~\cite{vbench} and EvalCrafter~\cite{evalcrafter} evaluate perceptual quality, while T2V-CompBench~\cite{t2vcompbench} and TC-Bench~\cite{tcbench} examine compositional and temporal constraints. VBench++~\cite{vbenchpp} and KeyFrame-Compass~\cite{keyframecompass} extend evaluation to image conditioning and keyframe adherence. StoryBench~\cite{storybench}, MSVBench~\cite{msvbench}, and LongAV-Compass~\cite{longavcompass} further address narrative progression and long-range consistency, with the latter covering text, image, and video conditioning. T2AV-Compass~\cite{t2avcompass} combines objective and subjective evaluation.
However, code-based video generation requires evaluating operational performance and the technical, semantic, and creative quality of complete videos.

\subsection{Code-Based Video and Animation Generation}

Prior work has explored code-based video and animation generation across specialized applications. Code2Video~\cite{code2video} introduces a Manim-based agent framework and the MMMC benchmark for educational videos, while DataReel~\cite{datareel} supports animated data storytelling. LogoMotion~\cite{logomotion} generates editable logo animations, and MG-Gen~\cite{mggen} produces layered HTML animations from images. InternSVG~\cite{internsvg} and LottieGPT~\cite{lottiegpt} explore structured vector animations.

Existing evaluations focus on specific tasks, with MMMC covering aesthetics, knowledge transfer, and efficiency~\cite{code2video}, Animation2Code~\cite{animation2code} measuring execution, appearance, and motion fidelity, MoVer~\cite{mover} verifying motion constraints, and HeyGen's Code2Video Benchmark~\cite{heygen} evaluating short promotional motion graphics. However, comprehensive evaluation of complete code-generated videos across diverse creative categories and input conditions remains limited.
Tab.~\ref{tab:benchmark_comparison} compares representative video generation benchmarks.

\section{Code2Video-Compass}
\label{sec:benchmark}

\subsection{Task Formulation}
\label{sec:task}
\benchmark evaluates general-purpose code-based video generation, where MLLMs translate textual instructions and optional visual references into executable projects. The rendering stack supports HTML/CSS, SVG, Canvas, and Three.js, with Remotion~\cite{remotion} producing the videos. Generated videos must exhibit coherent progression across events or visual states. Models reconstruct reference content through code and integrate it into the complete video. We consider three input settings:

\bpstart{Text-to-Video (T2V)}
Models interpret a self-contained instruction and generate code
to realize the specified visual content and temporal progression.

\bpstart{Image-to-Video (I2V)}
Models interpret the instruction and reference keyframe, understand its visual content, determine where it should appear in the complete video, and generate code to reconstruct and integrate its visual state.

\bpstart{Video-to-Video (V2V)}
Models interpret the instruction and reference clip, determine its placement in the complete video, and generate code to reconstruct and integrate its appearance, motion, and event sequence.

\subsection{Data Categories and Complexity}
\label{sec:categories}

We derive 10 categories from professional motion-design websites, award categories, template libraries, and creative portfolios~\cite{jittertemplates,artofthetitle,behancemotion}:
(1) Explainer Motion Graphics; (2) Branding \& Promotional Motion;
(3) Kinetic Typography; (4) UI \& Product Motion;
(5) Data \& Infographic Animation; (6) Title Sequences;
(7) Broadcast \& Show Graphics; (8) Illustrative \& Character Narrative;
(9) Abstract \& Experimental Motion; and (10) 3D Motion.
The categories are further refined into 24 subtypes using YouTube~\cite{youtube} examples, considering communication purpose, visual structure, and temporal organization. This taxonomy supports multiple labels per task and guides data collection and category-level analysis. We further define three complementary complexity dimensions, each with three levels.

\smallskip\noindent\textbf{Construction Complexity}
measures visual construction demands: standard primitives and component compositions (Level 1), custom geometry and detailed structures (Level 2), and procedural or simulation systems (Level 3).

\smallskip\noindent\textbf{Dynamic Animation Complexity}
measures temporal organization: staged sequences (Level 1), coordinated changes across elements and scenes (Level 2), and complex concurrent motion or transformations (Level 3).

\smallskip\noindent\textbf{State-Dependency Complexity}
measures the temporal span over which objects, data, and states must remain consistent: adjacent stages (Level 1), intervening stages (Level 2), and long-range dependencies across multiple changes (Level 3).

\subsection{Data Construction}
\label{sec:construction}

\begin{figure*}[t]
  \centering
  \captionsetup{skip=4pt}
  \includegraphics[width=\textwidth]{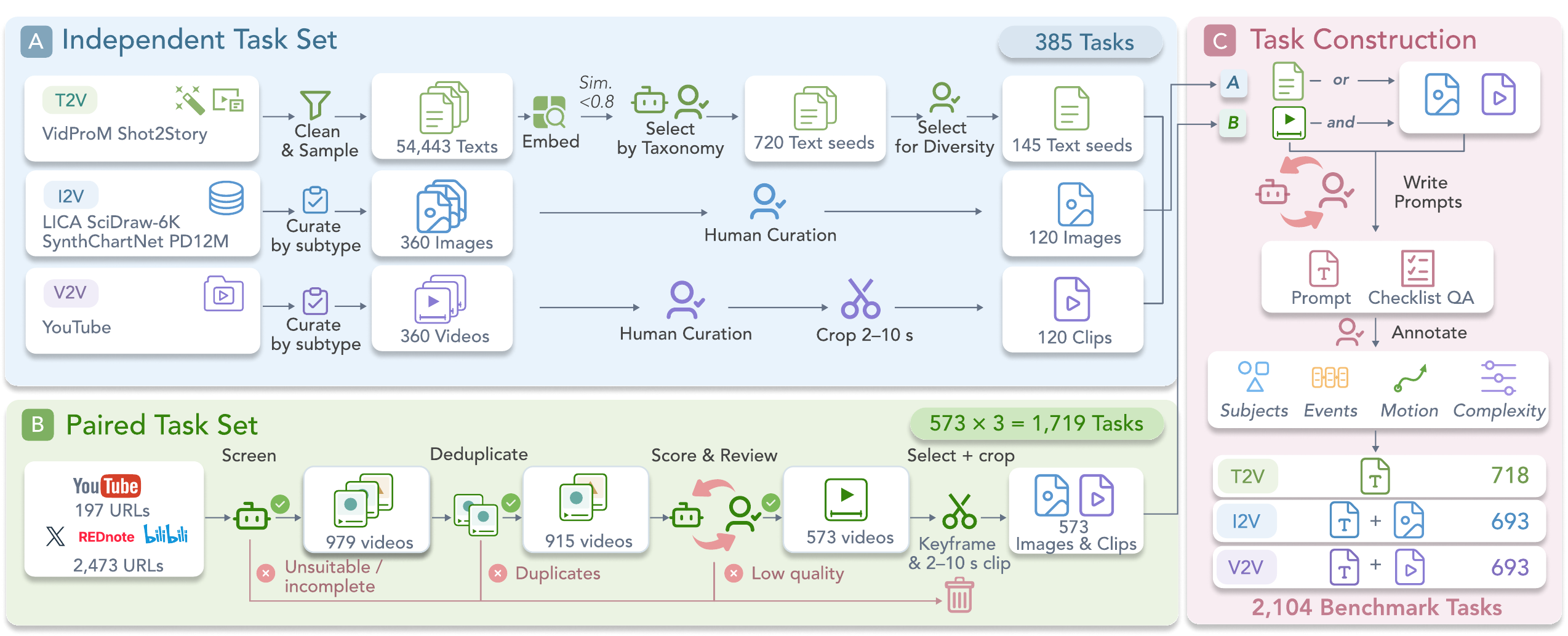}
  \caption{Data construction pipeline. (A) Independent tasks are developed from curated text seeds, reference images, and video clips. (B) Paired T2V, I2V, and V2V tasks are derived from shared complete source videos. (C) Both task sets undergo prompt construction, QA checklist development, and annotation of subjects, temporal events, motion types, and three complexity dimensions, yielding 2,104 tasks.}
  \label{fig:dataset_construction}
  \vspace{-15pt}
\end{figure*}

We construct 2,104 tasks through two complementary strategies, as illustrated in Fig.~\ref{fig:dataset_construction}. The Independent Task Set broadens creative coverage using text seeds and visual references, while the Paired Task Set grounds tasks in real-world creative practice by deriving them from complete professional and user-created videos.
Three co-authors conduct manual verification and quality review throughout the data construction process.

\bpstart{Independent Task Set}
For T2V, we source prompts from VidProM~\cite{vidprom} and video descriptions from Shot2Story~\cite{shot2story}. After source-specific cleaning, deterministic sampling yields 54,443 candidates, which we embed using all-MiniLM-L6-v2~\cite{minilm}. With GPT-6.1 Sol~\cite{gpt61sol} assistance and human review, taxonomy-guided curation narrows the pool to 720 candidates, from which we select 145 diverse task seeds.

For I2V, we curate reference images from LICA~\cite{lica}, SciDraw-6K~\cite{scidraw6k}, SynthChartNet~\cite{synthchartnet}, and PD12M~\cite{pd12m}. For V2V, we collect animation videos from YouTube~\cite{youtube}. For both settings, we curate 15 candidates per subtype and select five through content verification and manual quality review, yielding 120 references from 360 candidates in each setting. We crop a continuous 2--10\,s key segment from each selected video to form the V2V reference clip. These seeds and references guide complete-video task construction.

\bpstart{Paired Task Set}
To cover professional and user-created videos across platforms, we collect 197 animation records from YouTube~\cite{youtube} and 2,473 candidate URLs from Bilibili~\cite{bilibili}, REDnote~\cite{rednote}, and X~\cite{xplatform}. We screen candidates for video completeness and suitability for code-based generation with assistance from Gemini 3.8 Flash~\cite{gemini38flash}, yielding 979 videos. We remove exact duplicates using normalized URLs and SHA-256 hashes, and identify near-duplicates across all video sources using SSCD~\cite{sscd} frame embeddings indexed with FAISS~\cite{faiss}. Temporal matching distinguishes reposted works from videos with similar visual styles. After deduplication, we score the remaining 915 videos for narrative progression, content, visual design, and motion quality using Gemini 3.8 Flash. Human reviewers curate 573 complete videos, from which we select representative keyframes and crop continuous 2--10\,s reference clips with Gemini 3.8 Flash assistance and human verification. Each video yields aligned T2V, I2V, and V2V tasks with shared creative goals and event progression, totaling 1,719 tasks.

\bpstart{Prompt Construction and Annotation}
We develop detailed, self-contained instructions for both task sets with assistance from Gemini 3.8 Flash and human review. Independent instructions specify a new complete video based on each text seed or visual reference. For each paired source, we jointly develop T2V, I2V, and V2V instructions with shared core content requirements and event progression. Instructions specify visual elements, spatial relationships, exact text and data, motion, event sequences, and cross-stage dependencies, including reference content and surrounding events where applicable. We annotate subjects, temporal events, motion types, and three complexity dimensions, resolving ambiguities through reviewer discussion and consensus, and construct fine-grained QA checklists for instruction following, as described in Sec.~\ref{sec:subjective}. We iteratively review and refine these checklists with assistance from GPT-6.1 Sol.

\subsection{Dataset Statistics}
\label{sec:statistics}

\begin{figure*}[t]
  \centering
  \captionsetup{skip=4pt}
  \includegraphics[width=\textwidth]{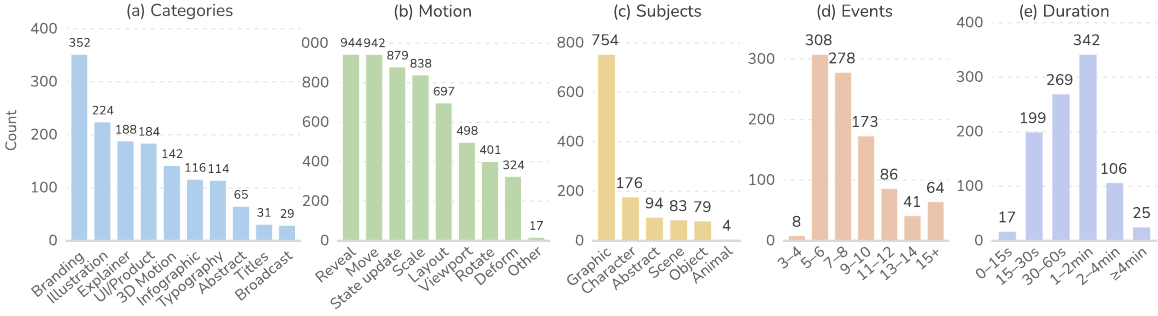}
  \caption{Distributions of (a) video categories, (b) motion types, (c) subject types, (d) event counts, and (e) target durations across 958 records (573 paired sources + 385 independent tasks). A record may have multiple video categories, motion types, and subject types.}
  \label{fig:dataset_distributions}
  \vspace{-10pt}
\end{figure*}

\begin{figure}[t]
  \centering
  \captionsetup{skip=4pt}
  \includegraphics[width=\linewidth]{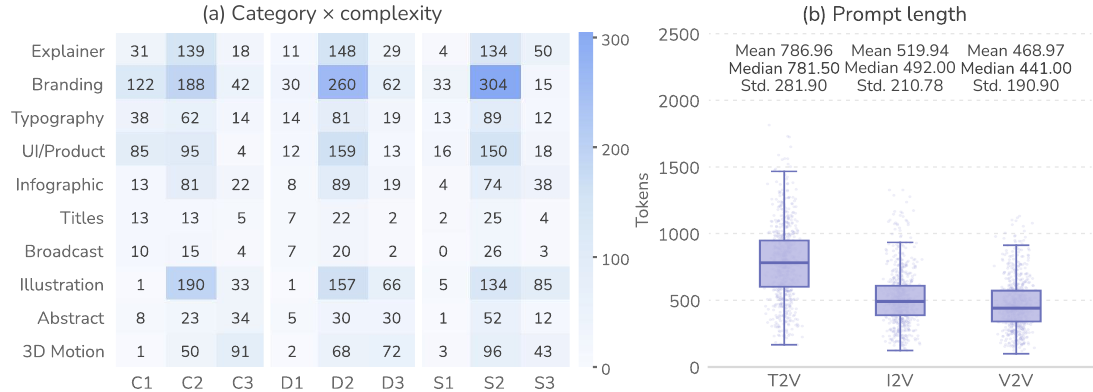}
  \caption{Task complexity and prompt length. C, D, and S denote construction, dynamic animation, and state-dependency complexity. (a) Record counts by category and complexity level (1--3), with multi-label records counted in each assigned category. (b) Prompt token distributions across T2V, I2V, and V2V, as in Tab.~\ref{tab:benchmark_comparison}.}
  \label{fig:dataset_complexity}
  \vspace{-15pt}
\end{figure}

The benchmark combines 1,719 paired tasks from 573 source videos with 385 independent tasks, totaling 2,104 tasks split into 718 T2V, 693 I2V, and 693 V2V tasks.
It covers required textual content in 19 languages (e.g., English, Greek, and Nepali) and includes 26,614 instruction-following checklist questions.
For coverage analysis, we count each source video and independent task once, yielding 958 records.
Figs.~\ref{fig:dataset_distributions} and \ref{fig:dataset_complexity} show the benchmark's breadth and diversity across creative categories, subjects, motion types, video duration, prompt length, and the three complexity dimensions.

\subsection{Dual-Level Evaluation Framework}
\label{sec:evaluation}

We introduce a dual-level evaluation framework comprising 12 objective metrics and five subjective metrics. Objective evaluation covers execution reliability, efficiency, video quality, text--video alignment, and reference fidelity. We employ Gemini 3.8 Flash~\cite{gemini38flash} to evaluate instruction following and rate four dimensions of holistic video quality.

\begingroup
\renewcommand{\bpstart}[1]{\par\addvspace{2pt}\noindent\mbox{\textbf{#1.}}\hspace{0.5em}\ignorespaces}

\subsubsection{Objective Evaluation}
\label{sec:objective}

\bpstart{\textcolor{HdrOperational}{Operational Performance}}
Practical video creation requires reliable execution and efficient generation. Execution success measures whether the code produces a usable video, while token consumption measures generation effort.

\metricstart{Execution Success (ES)}
Usable video projects must execute and render successfully. We report the fraction of sampled tasks producing complete, playable videos without errors under the shared backend and resource limits. Failed outputs receive zero on all non-operational metrics to prevent inflated averages from excluding difficult tasks.

\metricstart{Token Consumption (TC)}
We measure generation efficiency using mean API-reported output tokens per attempted task, consistent with token-based model comparisons such as Artificial Analysis~\cite{aamodelcomparison}.

\bpstart{\textcolor{HdrVideo}{Video Quality}}
Rendered videos need clear images, appealing presentation, and temporally consistent motion. The objective metrics capture these properties through technical quality, aesthetic quality, and local motion smoothness.

\metricstart{Technical Quality (TQ)}
Technical degradation, including blur, noise, and compression artifacts, affects the clarity of rendered content. The technical branch of DOVER++~\cite{dover} provides a score for this aspect of video quality.

\metricstart{Aesthetic Quality (AQ)}
The visual appeal of composition, color, and style contributes to effective presentation. Averaging frame-level scores from Aesthetic Predictor V2.5~\cite{aesthetic} gives a video-level estimate of aesthetic quality.

\metricstart{Motion Smoothness (MS)}
Smooth motion should be temporally predictable from neighboring frames. We resample videos to 24 FPS and use AMT~\cite{amt} to predict each intermediate frame from its two neighbors. The score is one minus the mean RGB absolute error normalized by 255, excluding comparisons across hard cuts.

\bpstart{\textcolor{HdrAlignment}{Text--Video Alignment}}
Instructions describe individual events and the overall progression of a complete video. Local TVAlign and Global TVAlign evaluate semantic correspondence at these two scales across T2V, I2V, and V2V.

\metricstart{Local TVAlign (L-TVA)}
Local TVAlign measures local semantic correspondence between eight-frame sliding video windows and text chunks derived from global and event-level descriptions. We compute window-level alignment scores using ViCLIP-L~\cite{internvid} and aggregate them across the video.

\metricstart{Global TVAlign (G-TVA)}
Global TVAlign measures semantic alignment between the complete video and the instruction as a whole, complementing local alignment and reference-fidelity metrics. We use Qwen3-VL-Embedding-8B~\cite{qwenembedding} to encode the instruction and frames spanning the video, then compute their embedding cosine similarity.

\bpstart{\textcolor{HdrImageFidelity}{I2V Reference Fidelity}}
Reference keyframes may appear at any stage, so we search the full generated video for the closest matching frame. Perceptual and structural fidelity measure visual resemblance and image structure.

\metricstart{Perceptual Fidelity (PF)}
Recognizable subjects and visual elements are central to reconstructing the reference image. We use DINOv3~\cite{dinov3} features to locate the closest matching frame and measure perceptual resemblance.

\metricstart{Structural Fidelity (SF)}
Faithful reconstruction also requires preserving the reference's local image structure. SSIM~\cite{ssim} measures luminance, contrast, and structural similarity at the same matching frame.

\bpstart{\textcolor{HdrClipFidelity}{V2V Reference Fidelity}}
The reference clip must be reconstructed within a complete video, so we first locate its corresponding interval in the output. Sequence alignment, appearance fidelity, and motion fidelity measure visual sequence correspondence, appearance similarity, and motion consistency within this interval.

\metricstart{Sequence Alignment (SA)}
To measure whether the reference's visual progression is preserved despite timing differences, we locate the corresponding interval using subsequence dynamic time warping (DTW) on DINOv3~\cite{dinov3} frame embeddings. The score is the mean cosine similarity along the matched path.

\metricstart{Appearance Fidelity (AF)}
The reconstructed interval should preserve the reference's subjects, colors, and shapes. Within the localized interval, we perform constrained DTW using DreamSim~\cite{dreamsim} cosine distances and define AF as one minus the path-averaged distance.

\metricstart{Motion Fidelity (MF)}
Similar visual appearance and temporal sequences do not necessarily imply similar motion patterns. We use CoTracker3~\cite{cotracker3} to extract point trajectories from the reference clip and the localized generated interval and compare their displacement directions and magnitudes after spatial and temporal normalization. We aggregate trajectory similarities through bidirectional best matching.

\subsubsection{Subjective Evaluation}
\label{sec:subjective}

\bpstart{\textcolor{HdrFollowing}{Instruction Following (IF)}}
We derive verifiable QA checklists from each instruction and employ Gemini 3.8 Flash~\cite{gemini38flash} to check whether the generated video satisfies each requirement. Each checklist contains up to 16 questions, including up to 14 generic questions across six dimensions and two task-specific questions. The six dimensions are (1) \textbf{Content}, covering exact text and information, (2) \textbf{Attribute}, covering appearance and quantity, (3) \textbf{Composition}, covering layout, framing, and color, (4) \textbf{Relations}, covering spatial and logical connections, (5) \textbf{Dynamics}, covering motion, interaction, transformation, viewport motion, and temporal relations, and (6) \textbf{Aesthetics}, covering style. The two task-specific questions cover full-video requirements and reference integration, including whether reference content appears at the required narrative stage and follows the specified event order. IF reports the fraction of questions passed.

\bpstart{\textcolor{HdrSubjective}{Holistic Video Quality}}
Creating an effective complete video requires coherent content, purposeful design, expressive motion, and sustained viewer interest. We rate these four aspects using five-point Likert scales.

\metricstart{Content Coherence (CC)}
Content coherence is rated based on progression, relationships among stages, state continuity, and consistency of text, data, charts, and interfaces.

\metricstart{Visual Design (VD)}
Visual design is rated based on composition, typography, readability, color, visual hierarchy, and purposeful detail within the intended style.

\metricstart{Motion Quality (MQ)}
Motion quality is rated based on easing, pacing, continuity, transitions, coordinated elements, temporal hierarchy, and behavioral variety, capturing both smoothness and expressive choreography.

\metricstart{Viewer Engagement (VE)}
Viewer engagement is rated by how well emphasis, pacing, and presentation sustain interest in a way that suits the video's intended use.

\endgroup

\section{Experiments}
\label{sec:experiments}

\subsection{Experimental Setup}

\noindent\textbf{Models.}
We evaluate 18 recent MLLMs across nine model families, including GPT~\cite{gpt6astra,gpt61sol,gpt6luna}, Claude~\cite{claudeopus55,claudesonnet55,claudehaiku55}, Gemini~\cite{gemini38flash,gemini4argon}, Qwen~\cite{qwen38max,qwen38flash,qwen3827b}, Kimi~\cite{kimik3,kimik31pro}, GLM~\cite{glm53flash}, DeepSeek~\cite{deepseekv41flash}, MiniMax~\cite{minimaxmodels,minimaxm31flash}, and MiMo~\cite{mimov26pro}.

\noindent\textbf{Evaluation Protocol.}
We sample approximately half the records from each of the two construction strategies, yielding a fixed subset of 1,054 tasks (360 T2V, 347 I2V, and 347 V2V) shared across all models.
The subset does not differ significantly from unsampled records in the distributions of the three complexity dimensions in Sec.~\ref{sec:categories} (Pearson's $\chi^2$ tests, $p_{\mathrm{Holm}}>0.05$, $V<0.02$) or prompt lengths (Welch's $t$-tests, $p_{\mathrm{Holm}}>0.05$, $|g|<0.05$).
Each model generates code from task instructions and applicable visual references in one API call using its provider's default reasoning settings, without execution feedback or repair.
Output token totals include API-reported reasoning tokens.
For V2V, GPT, Claude, and DeepSeek-V4.1-Flash receive 4-FPS frame sequences because their APIs do not accept native video input, while others receive native videos with sampling at 4 FPS.

\subsection{Main Results}
\label{sec:main_results}

\begin{table*}[!t]
  \centering
  \captionsetup{singlelinecheck=false,skip=7pt}

\caption{\textbf{Overall results across T2V, I2V, and V2V.}
Quality metrics aggregate all 1,054 tasks, while image and clip fidelity use the 347 I2V and 347 V2V tasks, respectively.
\textbf{Bold} and \underline{underlined} scores mark the best and second-best within each group, using unrounded values.}

  \label{tab:main_overall}
  \begingroup
  \footnotesize
  \setlength{\tabcolsep}{0.5pt}
  \renewcommand{\arraystretch}{1.15}
  \begin{tabular*}{\textwidth}{@{\extracolsep{\fill}}lc*{17}{c}@{}}
    \toprule
    \textbf{Model} & \textcolor{HdrOverall}{\textbf{Avg.}} & \multicolumn{2}{c}{\textcolor{HdrOperational}{\textbf{Operational}}} & \multicolumn{3}{c}{\textcolor{HdrVideo}{\textbf{Video Quality}}} & \multicolumn{2}{c}{\textcolor{HdrAlignment}{\textbf{Text Alignment}}} & \multicolumn{2}{c}{\textcolor{HdrImageFidelity}{\textbf{Image Fidelity}}} & \multicolumn{3}{c}{\textcolor{HdrClipFidelity}{\textbf{Clip Fidelity}}} & \textcolor{HdrFollowing}{\textbf{Instr. Follow.}}\hspace{1.5pt} & \multicolumn{4}{c}{\textcolor{HdrSubjective}{\textbf{Holistic Video Quality}}} \\
    \cmidrule(lr){3-4} \cmidrule(lr){5-7} \cmidrule(lr){8-9} \cmidrule(lr){10-11} \cmidrule(lr){12-14} \cmidrule(lr){15-15} \cmidrule(lr){16-19}
    & \textcolor{HdrOverall}{\textbf{Rank}$\downarrow$} & \textbf{ES}$\uparrow$ & \textbf{TC}$\downarrow$ & \textbf{TQ}$\uparrow$ & \textbf{AQ}$\uparrow$ & \textbf{MS}$\uparrow$ & \textbf{L-TVA}$\uparrow$ & \textbf{G-TVA}$\uparrow$ & \textbf{PF}$\uparrow$ & \textbf{SF}$\uparrow$ & \textbf{SA}$\uparrow$ & \textbf{AF}$\uparrow$ & \textbf{MF}$\uparrow$ & \textbf{IF}$\uparrow$ & \textbf{CC}$\uparrow$ & \textbf{VD}$\uparrow$ & \textbf{MQ}$\uparrow$ & \textbf{VE}$\uparrow$ \\
    \specialrule{\lightrulewidth}{\aboverulesep}{0pt}
    \rowcolor{ClosedSourceBg}\multicolumn{19}{@{}l}{\modelgroup{closedsourcecolor}{Proprietary Models}} \\
    \modelentry{\modelLogoClaude}{Claude Opus 5.5} & \avgrankcell{99.65}{\bestscore{1.067}} & \bestscore{0.958} & 26,159 & \secondscore{0.114} & \bestscore{3.727} & \bestscore{0.946} & \bestscore{0.522} & \bestscore{0.636} & \bestscore{0.894} & \bestscore{0.658} & \bestscore{0.882} & \bestscore{0.787} & \bestscore{0.445} & \bestscore{0.920} & \bestscore{3.993} & \bestscore{3.815} & \bestscore{3.807} & \bestscore{3.865} \\
    \modelentry{\modelLogoClaude}{Claude Sonnet 5.5} & \avgrankcell{95.17}{\secondscore{1.933}} & \secondscore{0.951} & 37,460 & \bestscore{0.116} & \secondscore{3.681} & \secondscore{0.938} & \secondscore{0.521} & \secondscore{0.608} & \secondscore{0.890} & \secondscore{0.657} & \secondscore{0.873} & \secondscore{0.773} & \secondscore{0.443} & \secondscore{0.857} & \secondscore{3.753} & \secondscore{3.717} & \secondscore{3.681} & \secondscore{3.698} \\
    \modelentry{\modelLogoOpenAI}{GPT-6 Astra} & \avgrankcell{81.88}{4.500} & 0.862 & 13,867 & 0.110 & 3.536 & 0.854 & 0.478 & 0.575 & 0.834 & 0.649 & 0.790 & 0.727 & 0.401 & 0.842 & 3.677 & 3.641 & 3.539 & 3.607 \\
    \modelentry{\modelLogoGemini}{Gemini 4 Argon} & \avgrankcell{81.36}{4.600} & 0.921 & 16,411 & 0.107 & 3.510 & 0.910 & 0.501 & 0.595 & 0.874 & 0.635 & 0.783 & 0.677 & 0.403 & 0.788 & 3.545 & 3.524 & 3.437 & 3.496 \\
    \modelentry{\modelLogoOpenAI}{GPT-6.1 Sol} & \avgrankcell{78.09}{5.233} & 0.870 & \secondscore{10,993} & 0.111 & 3.524 & 0.861 & 0.482 & 0.578 & 0.828 & 0.635 & 0.779 & 0.710 & 0.386 & 0.840 & 3.676 & 3.631 & 3.527 & 3.607 \\
    \modelentry{\modelLogoClaude}{Claude Haiku 5.5} & \avgrankcell{70.32}{6.733} & 0.895 & 40,549 & 0.104 & 3.392 & 0.886 & 0.481 & 0.579 & 0.857 & 0.629 & 0.801 & 0.668 & 0.393 & 0.790 & 3.324 & 3.145 & 3.118 & 3.175 \\
    \modelentry{\modelLogoGemini}{Gemini 3.8 Flash} & \avgrankcell{69.98}{6.800} & 0.919 & 15,541 & 0.102 & 3.476 & 0.911 & 0.491 & 0.595 & 0.839 & 0.586 & 0.781 & 0.642 & 0.393 & 0.769 & 3.291 & 3.301 & 3.124 & 3.233 \\
    \modelentry{\modelLogoOpenAI}{GPT-6 Luna} & \avgrankcell{60.66}{8.600} & 0.897 & \bestscore{8,619} & 0.109 & 3.541 & 0.890 & 0.487 & 0.565 & 0.823 & 0.617 & 0.782 & 0.661 & 0.383 & 0.684 & 2.868 & 3.046 & 2.721 & 2.841 \\
    \modelentry{\modelLogoQwen}{Qwen3.8-Max} & \avgrankcell{32.02}{14.133} & 0.741 & 48,646 & 0.085 & 2.817 & 0.734 & 0.400 & 0.489 & 0.587 & 0.424 & 0.616 & 0.533 & 0.317 & 0.684 & 2.924 & 2.748 & 2.768 & 2.805 \\
    \modelentry{\modelLogoQwen}{Qwen3.8-Flash} & \avgrankcell{27.87}{14.933} & 0.739 & 49,246 & 0.085 & 2.805 & 0.732 & 0.394 & 0.485 & 0.694 & 0.469 & 0.607 & 0.504 & 0.306 & 0.658 & 2.721 & 2.562 & 2.559 & 2.595 \\
    \specialrule{\lightrulewidth}{\aboverulesep}{0pt}
    \rowcolor{OpenWeightBg}\multicolumn{19}{@{}l}{\modelgroup{openweightcolor}{Open-weight Models}} \\
    \modelentry{\modelLogoMiMo}{MiMo-V2.6-Pro} & \avgrankcell{59.97}{\bestscore{8.733}} & \bestscore{0.870} & \secondscore{28,769} & \bestscore{0.108} & \bestscore{3.366} & \bestscore{0.862} & \bestscore{0.473} & \bestscore{0.570} & \bestscore{0.788} & \secondscore{0.540} & \bestscore{0.757} & \bestscore{0.624} & \bestscore{0.374} & 0.767 & 3.273 & 3.194 & 3.130 & 3.197 \\
    \modelentry{\modelLogoKimi}{Kimi K3.1 Pro} & \avgrankcell{58.24}{\secondscore{9.067}} & 0.817 & 30,092 & 0.099 & 3.225 & 0.809 & \secondscore{0.449} & 0.542 & 0.743 & \bestscore{0.542} & \secondscore{0.734} & \secondscore{0.615} & \secondscore{0.372} & \bestscore{0.774} & \bestscore{3.485} & \bestscore{3.419} & \bestscore{3.369} & \bestscore{3.425} \\
    \modelentry{\modelLogoDeepSeek}{DeepSeek-V4.1-Flash} & \avgrankcell{55.14}{9.667} & \secondscore{0.834} & 64,159 & \secondscore{0.100} & \secondscore{3.237} & \secondscore{0.826} & 0.448 & \secondscore{0.551} & 0.746 & 0.503 & 0.713 & 0.598 & 0.354 & \secondscore{0.774} & \secondscore{3.324} & \secondscore{3.231} & \secondscore{3.198} & \secondscore{3.240} \\
    \modelentry{\modelLogoKimi}{Kimi K3} & \avgrankcell{44.09}{11.800} & 0.815 & 39,135 & 0.093 & 3.157 & 0.807 & 0.436 & 0.538 & \secondscore{0.776} & 0.539 & 0.639 & 0.506 & 0.321 & 0.734 & 3.114 & 3.004 & 2.947 & 2.991 \\
    \modelentry{\modelLogoGLM}{GLM-5.3-Flash} & \avgrankcell{40.30}{12.533} & 0.742 & 90,005 & 0.087 & 2.898 & 0.735 & 0.402 & 0.490 & 0.701 & 0.485 & 0.684 & 0.569 & 0.345 & 0.696 & 3.004 & 2.925 & 2.891 & 2.927 \\
    \modelentry{\modelLogoQwen}{Qwen3.8-27B} & \avgrankcell{23.39}{15.800} & 0.705 & 62,086 & 0.080 & 2.717 & 0.699 & 0.374 & 0.460 & 0.646 & 0.452 & 0.604 & 0.506 & 0.303 & 0.632 & 2.639 & 2.534 & 2.502 & 2.536 \\
    \modelentry{\modelLogoMiniMax}{MiniMax M3.1 Flash} & \avgrankcell{17.87}{16.867} & 0.646 & 41,370 & 0.076 & 2.500 & 0.641 & 0.348 & 0.419 & 0.617 & 0.425 & 0.582 & 0.489 & 0.291 & 0.567 & 2.396 & 2.379 & 2.324 & 2.367 \\
    \modelentry{\modelLogoMiniMax}{MiniMax M3} & \avgrankcell{12.00}{18.000} & 0.597 & \bestscore{13,167} & 0.059 & 2.232 & 0.593 & 0.297 & 0.370 & 0.491 & 0.348 & 0.544 & 0.410 & 0.280 & 0.404 & 1.605 & 1.604 & 1.500 & 1.528 \\
    \bottomrule
  \end{tabular*}
  \endgroup
\end{table*}

\begin{table*}[!tp]
  \centering
  \captionsetup{singlelinecheck=false,skip=7pt}
  \caption{\textbf{Detailed results on T2V.}
  We average ranks across the applicable metrics in Sec.~\ref{sec:evaluation}.
  Scoring and notation follow Tab.~\ref{tab:main_overall}.}
  \label{tab:main_t2v}
  \begingroup
  \footnotesize
  \setlength{\tabcolsep}{2pt}
  \renewcommand{\arraystretch}{1.15}
  \begin{tabular*}{\textwidth}{@{\extracolsep{\fill}}lc*{12}{c}@{}}
    \toprule
    \textbf{Model} & \textcolor{HdrOverall}{\textbf{Avg.}} & \multicolumn{2}{c}{\textcolor{HdrOperational}{\textbf{Operational}}} & \multicolumn{3}{c}{\textcolor{HdrVideo}{\textbf{Video Quality}}} & \multicolumn{2}{c}{\textcolor{HdrAlignment}{\textbf{Text Alignment}}} & \textcolor{HdrFollowing}{\textbf{Instr. Follow.}} & \multicolumn{4}{c}{\textcolor{HdrSubjective}{\textbf{Holistic Video Quality}}} \\
    \cmidrule(lr){3-4} \cmidrule(lr){5-7} \cmidrule(lr){8-9} \cmidrule(lr){10-10} \cmidrule(lr){11-14}
    & \textcolor{HdrOverall}{\textbf{Rank}$\downarrow$} & \textbf{ES}$\uparrow$ & \textbf{TC}$\downarrow$ & \textbf{TQ}$\uparrow$ & \textbf{AQ}$\uparrow$ & \textbf{MS}$\uparrow$ & \textbf{L-TVA}$\uparrow$ & \textbf{G-TVA}$\uparrow$ & \textbf{IF}$\uparrow$ & \textbf{CC}$\uparrow$ & \textbf{VD}$\uparrow$ & \textbf{MQ}$\uparrow$ & \textbf{VE}$\uparrow$ \\
    \specialrule{\lightrulewidth}{\aboverulesep}{0pt}
    \rowcolor{ClosedSourceBg}\multicolumn{14}{@{}l}{\modelgroup{closedsourcecolor}{Proprietary Models}} \\
    \modelentry{\modelLogoClaude}{Claude Opus 5.5} & \avgrankcell{96.89}{\bestscore{1.600}} & \bestscore{0.944} & 29,400 & 0.115 & \bestscore{3.636} & \bestscore{0.933} & \bestscore{0.522} & \bestscore{0.653} & \bestscore{0.915} & \bestscore{4.047} & 3.808 & \secondscore{3.828} & \secondscore{3.894} \\
    \modelentry{\modelLogoClaude}{Claude Sonnet 5.5} & \avgrankcell{95.34}{\secondscore{1.900}} & \secondscore{0.936} & 48,679 & \secondscore{0.116} & 3.587 & \secondscore{0.924} & \secondscore{0.521} & \secondscore{0.648} & \secondscore{0.911} & \secondscore{4.039} & \secondscore{3.861} & \bestscore{3.847} & \bestscore{3.900} \\
    \modelentry{\modelLogoOpenAI}{GPT-6.1 Sol} & \avgrankcell{91.20}{2.700} & 0.892 & \secondscore{13,022} & \bestscore{0.117} & \secondscore{3.631} & 0.884 & 0.501 & 0.614 & 0.873 & 3.953 & \bestscore{3.883} & 3.767 & 3.861 \\
    \modelentry{\modelLogoOpenAI}{GPT-6 Astra} & \avgrankcell{79.29}{5.000} & 0.864 & 16,647 & 0.112 & 3.561 & 0.856 & 0.482 & 0.594 & 0.850 & 3.806 & 3.756 & 3.703 & 3.747 \\
    \modelentry{\modelLogoGemini}{Gemini 4 Argon} & \avgrankcell{75.67}{5.700} & 0.908 & 16,225 & 0.106 & 3.405 & 0.898 & 0.495 & 0.611 & 0.772 & 3.486 & 3.442 & 3.367 & 3.442 \\
    \modelentry{\modelLogoGemini}{Gemini 3.8 Flash} & \avgrankcell{69.46}{6.900} & 0.903 & 17,817 & 0.100 & 3.332 & 0.894 & 0.482 & 0.614 & 0.798 & 3.481 & 3.383 & 3.272 & 3.375 \\
    \modelentry{\modelLogoQwen}{Qwen3.8-Max} & \avgrankcell{67.13}{7.350} & 0.875 & 60,316 & 0.105 & 3.271 & 0.867 & 0.481 & 0.598 & 0.812 & 3.561 & 3.253 & 3.308 & 3.356 \\
    \modelentry{\modelLogoOpenAI}{GPT-6 Luna} & \avgrankcell{60.14}{8.700} & 0.883 & \bestscore{9,923} & 0.109 & 3.461 & 0.877 & 0.483 & 0.600 & 0.732 & 3.111 & 3.200 & 2.908 & 3.072 \\
    \modelentry{\modelLogoClaude}{Claude Haiku 5.5} & \avgrankcell{40.47}{12.500} & 0.808 & 47,634 & 0.096 & 2.989 & 0.800 & 0.445 & 0.551 & 0.737 & 3.161 & 2.872 & 2.889 & 2.950 \\
    \modelentry{\modelLogoQwen}{Qwen3.8-Flash} & \avgrankcell{28.56}{14.800} & 0.739 & 65,945 & 0.088 & 2.763 & 0.732 & 0.400 & 0.500 & 0.641 & 2.639 & 2.469 & 2.481 & 2.492 \\
    \specialrule{\lightrulewidth}{\aboverulesep}{0pt}
    \rowcolor{OpenWeightBg}\multicolumn{14}{@{}l}{\modelgroup{openweightcolor}{Open-weight Models}} \\
    \modelentry{\modelLogoDeepSeek}{DeepSeek-V4.1-Flash} & \avgrankcell{60.40}{\bestscore{8.650}} & \secondscore{0.853} & 65,680 & \secondscore{0.104} & \bestscore{3.276} & \secondscore{0.845} & \secondscore{0.462} & \bestscore{0.585} & \bestscore{0.797} & \secondscore{3.469} & \secondscore{3.303} & \secondscore{3.308} & \secondscore{3.353} \\
    \modelentry{\modelLogoKimi}{Kimi K3.1 Pro} & \avgrankcell{59.11}{\secondscore{8.900}} & 0.806 & 33,262 & 0.099 & 3.178 & 0.798 & 0.445 & 0.551 & \secondscore{0.774} & \bestscore{3.494} & \bestscore{3.433} & \bestscore{3.375} & \bestscore{3.433} \\
    \modelentry{\modelLogoMiMo}{MiMo-V2.6-Pro} & \avgrankcell{53.41}{10.000} & \bestscore{0.861} & \secondscore{22,741} & \bestscore{0.108} & \secondscore{3.264} & \bestscore{0.854} & \bestscore{0.473} & \secondscore{0.585} & 0.740 & 3.189 & 3.089 & 3.036 & 3.094 \\
    \modelentry{\modelLogoKimi}{Kimi K3} & \avgrankcell{47.72}{11.100} & 0.819 & 39,681 & 0.094 & 3.120 & 0.812 & 0.443 & 0.562 & 0.756 & 3.247 & 3.100 & 3.064 & 3.125 \\
    \modelentry{\modelLogoGLM}{GLM-5.3-Flash} & \avgrankcell{28.05}{14.900} & 0.678 & 94,000 & 0.081 & 2.623 & 0.671 & 0.372 & 0.463 & 0.640 & 2.831 & 2.719 & 2.683 & 2.725 \\
    \modelentry{\modelLogoQwen}{Qwen3.8-27B} & \avgrankcell{25.98}{15.300} & 0.692 & 87,862 & 0.079 & 2.676 & 0.686 & 0.368 & 0.464 & 0.621 & 2.669 & 2.547 & 2.531 & 2.567 \\
    \modelentry{\modelLogoMiniMax}{MiniMax M3.1 Flash} & \avgrankcell{17.18}{17.000} & 0.592 & 42,188 & 0.072 & 2.298 & 0.587 & 0.322 & 0.398 & 0.532 & 2.264 & 2.233 & 2.194 & 2.231 \\
    \modelentry{\modelLogoMiniMax}{MiniMax M3} & \avgrankcell{12.00}{18.000} & 0.519 & \bestscore{15,423} & 0.052 & 1.930 & 0.516 & 0.262 & 0.332 & 0.330 & 1.356 & 1.344 & 1.258 & 1.294 \\
    \bottomrule
  \end{tabular*}
  \endgroup
\end{table*}

\begin{table*}[!tp]
  \centering
  \captionsetup{singlelinecheck=false,skip=7pt}
  \caption{\textbf{Detailed results on I2V.}
  Score and rank conventions follow Tab.~\ref{tab:main_t2v}.}
  \label{tab:main_i2v}
  \begingroup
  \footnotesize
  \setlength{\tabcolsep}{2pt}
  \renewcommand{\arraystretch}{1.15}
  \begin{tabular*}{\textwidth}{@{\extracolsep{\fill}}lc*{14}{c}@{}}
    \toprule
    \textbf{Model} & \textcolor{HdrOverall}{\textbf{Avg.}} & \multicolumn{2}{c}{\textcolor{HdrOperational}{\textbf{Operational}}} & \multicolumn{3}{c}{\textcolor{HdrVideo}{\textbf{Video Quality}}} & \multicolumn{2}{c}{\textcolor{HdrAlignment}{\textbf{Text Alignment}}} & \multicolumn{2}{c}{\textcolor{HdrImageFidelity}{\textbf{Image Fidelity}}} & \textcolor{HdrFollowing}{\textbf{Instr. Follow.}} & \multicolumn{4}{c}{\textcolor{HdrSubjective}{\textbf{Holistic Video Quality}}} \\
    \cmidrule(lr){3-4} \cmidrule(lr){5-7} \cmidrule(lr){8-9} \cmidrule(lr){10-11} \cmidrule(lr){12-12} \cmidrule(lr){13-16}
    & \textcolor{HdrOverall}{\textbf{Rank}$\downarrow$} & \textbf{ES}$\uparrow$ & \textbf{TC}$\downarrow$ & \textbf{TQ}$\uparrow$ & \textbf{AQ}$\uparrow$ & \textbf{MS}$\uparrow$ & \textbf{L-TVA}$\uparrow$ & \textbf{G-TVA}$\uparrow$ & \textbf{PF}$\uparrow$ & \textbf{SF}$\uparrow$ & \textbf{IF}$\uparrow$ & \textbf{CC}$\uparrow$ & \textbf{VD}$\uparrow$ & \textbf{MQ}$\uparrow$ & \textbf{VE}$\uparrow$ \\
    \specialrule{\lightrulewidth}{\aboverulesep}{0pt}
    \rowcolor{ClosedSourceBg}\multicolumn{16}{@{}l}{\modelgroup{closedsourcecolor}{Proprietary Models}} \\
    \modelentry{\modelLogoClaude}{Claude Opus 5.5} & \avgrankcell{99.14}{\bestscore{1.167}} & \bestscore{0.960} & 19,905 & \secondscore{0.112} & \bestscore{3.754} & \bestscore{0.948} & \secondscore{0.517} & \bestscore{0.625} & \bestscore{0.894} & \bestscore{0.658} & \bestscore{0.915} & \bestscore{3.934} & \bestscore{3.784} & \bestscore{3.755} & \bestscore{3.818} \\
    \modelentry{\modelLogoClaude}{Claude Sonnet 5.5} & \avgrankcell{94.82}{\secondscore{2.000}} & \secondscore{0.954} & 28,926 & \bestscore{0.114} & \secondscore{3.716} & \secondscore{0.942} & 0.514 & \secondscore{0.620} & \secondscore{0.890} & \secondscore{0.657} & \secondscore{0.901} & \secondscore{3.853} & \secondscore{3.718} & \secondscore{3.683} & \secondscore{3.749} \\
    \modelentry{\modelLogoGemini}{Gemini 4 Argon} & \avgrankcell{85.76}{3.750} & 0.948 & 16,381 & 0.111 & 3.666 & 0.937 & \bestscore{0.521} & 0.607 & 0.874 & 0.635 & 0.822 & 3.620 & 3.620 & 3.470 & 3.542 \\
    \modelentry{\modelLogoOpenAI}{GPT-6 Astra} & \avgrankcell{81.88}{4.500} & 0.873 & 12,689 & 0.111 & 3.603 & 0.866 & 0.481 & 0.574 & 0.834 & 0.649 & 0.855 & 3.692 & 3.657 & 3.467 & 3.591 \\
    \modelentry{\modelLogoClaude}{Claude Haiku 5.5} & \avgrankcell{78.43}{5.167} & 0.948 & 35,606 & 0.111 & 3.665 & 0.938 & 0.505 & 0.605 & 0.857 & 0.629 & 0.838 & 3.493 & 3.378 & 3.288 & 3.380 \\
    \modelentry{\modelLogoOpenAI}{GPT-6.1 Sol} & \avgrankcell{73.25}{6.167} & 0.870 & \secondscore{10,015} & 0.110 & 3.539 & 0.863 & 0.478 & 0.570 & 0.828 & 0.635 & 0.837 & 3.631 & 3.548 & 3.435 & 3.550 \\
    \modelentry{\modelLogoGemini}{Gemini 3.8 Flash} & \avgrankcell{62.69}{8.208} & 0.931 & 15,014 & 0.103 & 3.577 & 0.922 & 0.499 & 0.596 & 0.839 & 0.586 & 0.745 & 3.118 & 3.210 & 2.888 & 3.055 \\
    \modelentry{\modelLogoOpenAI}{GPT-6 Luna} & \avgrankcell{55.57}{9.583} & 0.902 & \bestscore{8,833} & 0.108 & 3.556 & 0.896 & 0.490 & 0.579 & 0.823 & 0.617 & 0.703 & 2.882 & 3.075 & 2.611 & 2.793 \\
    \modelentry{\modelLogoQwen}{Qwen3.8-Flash} & \avgrankcell{35.29}{13.500} & 0.769 & 54,246 & 0.087 & 2.938 & 0.763 & 0.406 & 0.499 & 0.694 & 0.469 & 0.704 & 2.968 & 2.807 & 2.769 & 2.841 \\
    \modelentry{\modelLogoQwen}{Qwen3.8-Max} & \avgrankcell{17.18}{17.000} & 0.640 & 42,967 & 0.072 & 2.479 & 0.634 & 0.341 & 0.413 & 0.587 & 0.424 & 0.598 & 2.516 & 2.427 & 2.418 & 2.458 \\
    \specialrule{\lightrulewidth}{\aboverulesep}{0pt}
    \rowcolor{OpenWeightBg}\multicolumn{16}{@{}l}{\modelgroup{openweightcolor}{Open-weight Models}} \\
    \modelentry{\modelLogoMiMo}{MiMo-V2.6-Pro} & \avgrankcell{59.02}{\bestscore{8.917}} & \bestscore{0.870} & \secondscore{27,446} & \bestscore{0.108} & \bestscore{3.404} & \bestscore{0.863} & \bestscore{0.473} & \bestscore{0.568} & \bestscore{0.788} & \secondscore{0.540} & \secondscore{0.778} & 3.282 & 3.233 & 3.121 & 3.199 \\
    \modelentry{\modelLogoKimi}{Kimi K3} & \avgrankcell{56.00}{\secondscore{9.500}} & \secondscore{0.865} & 39,300 & 0.099 & \secondscore{3.391} & \secondscore{0.856} & \secondscore{0.462} & \secondscore{0.563} & \secondscore{0.776} & 0.539 & \bestscore{0.783} & \secondscore{3.337} & \secondscore{3.251} & 3.138 & 3.196 \\
    \modelentry{\modelLogoKimi}{Kimi K3.1 Pro} & \avgrankcell{55.57}{9.583} & 0.795 & 31,447 & 0.096 & 3.131 & 0.788 & 0.436 & 0.521 & 0.743 & \bestscore{0.542} & 0.756 & \bestscore{3.504} & \bestscore{3.432} & \bestscore{3.380} & \bestscore{3.452} \\
    \modelentry{\modelLogoDeepSeek}{DeepSeek-V4.1-Flash} & \avgrankcell{52.33}{10.208} & 0.824 & 61,171 & \secondscore{0.099} & 3.222 & 0.817 & 0.442 & 0.536 & 0.746 & 0.503 & 0.766 & 3.280 & 3.210 & \secondscore{3.141} & \secondscore{3.210} \\
    \modelentry{\modelLogoGLM}{GLM-5.3-Flash} & \avgrankcell{39.18}{12.750} & 0.767 & 88,070 & 0.089 & 3.014 & 0.759 & 0.411 & 0.498 & 0.701 & 0.485 & 0.715 & 3.066 & 2.977 & 2.939 & 2.988 \\
    \modelentry{\modelLogoQwen}{Qwen3.8-27B} & \avgrankcell{27.53}{15.000} & 0.718 & 50,307 & 0.081 & 2.798 & 0.712 & 0.378 & 0.465 & 0.646 & 0.452 & 0.651 & 2.651 & 2.559 & 2.481 & 2.542 \\
    \modelentry{\modelLogoMiniMax}{MiniMax M3.1 Flash} & \avgrankcell{22.35}{16.000} & 0.671 & 38,089 & 0.079 & 2.608 & 0.666 & 0.361 & 0.434 & 0.617 & 0.425 & 0.603 & 2.539 & 2.519 & 2.435 & 2.493 \\
    \modelentry{\modelLogoMiniMax}{MiniMax M3} & \avgrankcell{12.00}{18.000} & 0.588 & \bestscore{12,115} & 0.059 & 2.219 & 0.584 & 0.289 & 0.360 & 0.491 & 0.348 & 0.403 & 1.591 & 1.605 & 1.476 & 1.513 \\
    \bottomrule
  \end{tabular*}
  \endgroup
\end{table*}

\begin{table*}[!tp]
  \centering
  \captionsetup{singlelinecheck=false,skip=7pt}
  \caption{\textbf{Detailed results on V2V.}
  Models marked \frameinput{} use frames and all others use native video.
  Score and rank conventions follow Tab.~\ref{tab:main_t2v}.}
  \label{tab:main_v2v}
  \begingroup
  \footnotesize
  \setlength{\tabcolsep}{1.6pt}
  \renewcommand{\arraystretch}{1.15}
  \begin{tabular*}{\textwidth}{@{\extracolsep{\fill}}lc*{15}{c}@{}}
    \toprule
    \textbf{Model} & \textcolor{HdrOverall}{\textbf{Avg.}} & \multicolumn{2}{c}{\textcolor{HdrOperational}{\textbf{Operational}}} & \multicolumn{3}{c}{\textcolor{HdrVideo}{\textbf{Video Quality}}} & \multicolumn{2}{c}{\textcolor{HdrAlignment}{\textbf{Text Alignment}}} & \multicolumn{3}{c}{\textcolor{HdrClipFidelity}{\textbf{Clip Fidelity}}} & \textcolor{HdrFollowing}{\textbf{Instr. Follow.}} & \multicolumn{4}{c}{\textcolor{HdrSubjective}{\textbf{Holistic Video Quality}}} \\
    \cmidrule(lr){3-4} \cmidrule(lr){5-7} \cmidrule(lr){8-9} \cmidrule(lr){10-12} \cmidrule(lr){13-13} \cmidrule(lr){14-17}
    & \textcolor{HdrOverall}{\textbf{Rank}$\downarrow$} & \textbf{ES}$\uparrow$ & \textbf{TC}$\downarrow$ & \textbf{TQ}$\uparrow$ & \textbf{AQ}$\uparrow$ & \textbf{MS}$\uparrow$ & \textbf{L-TVA}$\uparrow$ & \textbf{G-TVA}$\uparrow$ & \textbf{SA}$\uparrow$ & \textbf{AF}$\uparrow$ & \textbf{MF}$\uparrow$ & \textbf{IF}$\uparrow$ & \textbf{CC}$\uparrow$ & \textbf{VD}$\uparrow$ & \textbf{MQ}$\uparrow$ & \textbf{VE}$\uparrow$ \\
    \specialrule{\lightrulewidth}{\aboverulesep}{0pt}
    \rowcolor{ClosedSourceBg}\multicolumn{17}{@{}l}{\modelgroup{closedsourcecolor}{Proprietary Models}} \\
    \modelentry{\modelLogoClaude}{Claude Opus 5.5}\nobreak\hspace{2pt}\frameinput & \avgrankcell{99.20}{\bestscore{1.154}} & \bestscore{0.971} & 29,050 & \secondscore{0.115} & \bestscore{3.794} & \bestscore{0.958} & \secondscore{0.527} & \bestscore{0.630} & \bestscore{0.882} & \bestscore{0.787} & \bestscore{0.445} & \bestscore{0.929} & \bestscore{3.997} & \bestscore{3.853} & \bestscore{3.839} & \bestscore{3.882} \\
    \modelentry{\modelLogoClaude}{Claude Sonnet 5.5}\nobreak\hspace{2pt}\frameinput & \avgrankcell{88.05}{\secondscore{3.308}} & \secondscore{0.963} & 34,355 & \bestscore{0.118} & \secondscore{3.744} & \secondscore{0.948} & \bestscore{0.528} & 0.555 & \secondscore{0.873} & \secondscore{0.773} & \secondscore{0.443} & 0.756 & 3.357 & \secondscore{3.568} & \secondscore{3.507} & 3.438 \\
    \modelentry{\modelLogoGemini}{Gemini 4 Argon} & \avgrankcell{81.09}{4.654} & 0.908 & 16,635 & 0.104 & 3.463 & 0.897 & 0.487 & 0.567 & 0.783 & 0.677 & 0.403 & 0.770 & \secondscore{3.530} & 3.513 & 3.478 & \secondscore{3.507} \\
    \modelentry{\modelLogoOpenAI}{GPT-6 Astra}\nobreak\hspace{2pt}\frameinput & \avgrankcell{80.69}{4.731} & 0.850 & 12,158 & 0.107 & 3.445 & 0.840 & 0.471 & 0.557 & 0.790 & 0.727 & 0.401 & \secondscore{0.821} & \secondscore{3.530} & 3.507 & 3.441 & 3.478 \\
    \modelentry{\modelLogoClaude}{Claude Haiku 5.5}\nobreak\hspace{2pt}\frameinput & \avgrankcell{76.11}{5.615} & 0.931 & 38,142 & 0.107 & 3.538 & 0.921 & 0.495 & \secondscore{0.580} & 0.801 & 0.668 & 0.393 & 0.797 & 3.326 & 3.196 & 3.184 & 3.202 \\
    \modelentry{\modelLogoGemini}{Gemini 3.8 Flash} & \avgrankcell{70.93}{6.615} & 0.925 & 13,708 & 0.102 & 3.526 & 0.916 & 0.492 & 0.574 & 0.781 & 0.642 & 0.393 & 0.764 & 3.268 & 3.305 & 3.207 & 3.265 \\
    \modelentry{\modelLogoOpenAI}{GPT-6.1 Sol}\nobreak\hspace{2pt}\frameinput & \avgrankcell{70.93}{6.615} & 0.847 & \secondscore{9,862} & 0.107 & 3.399 & 0.837 & 0.466 & 0.549 & 0.779 & 0.710 & 0.386 & 0.808 & 3.432 & 3.452 & 3.369 & 3.401 \\
    \modelentry{\modelLogoOpenAI}{GPT-6 Luna}\nobreak\hspace{2pt}\frameinput & \avgrankcell{59.58}{8.808} & 0.905 & \bestscore{7,050} & 0.110 & 3.608 & 0.899 & 0.488 & 0.514 & 0.782 & 0.661 & 0.383 & 0.614 & 2.602 & 2.856 & 2.637 & 2.648 \\
    \modelentry{\modelLogoQwen}{Qwen3.8-Max} & \avgrankcell{30.71}{14.385} & 0.703 & 42,217 & 0.078 & 2.685 & 0.696 & 0.376 & 0.452 & 0.616 & 0.533 & 0.317 & 0.637 & 2.671 & 2.545 & 2.556 & 2.579 \\
    \modelentry{\modelLogoQwen}{Qwen3.8-Flash} & \avgrankcell{27.13}{15.077} & 0.709 & 26,971 & 0.079 & 2.715 & 0.702 & 0.375 & 0.454 & 0.607 & 0.504 & 0.306 & 0.628 & 2.559 & 2.412 & 2.429 & 2.455 \\
    \specialrule{\lightrulewidth}{\aboverulesep}{0pt}
    \rowcolor{OpenWeightBg}\multicolumn{17}{@{}l}{\modelgroup{openweightcolor}{Open-weight Models}} \\
    \modelentry{\modelLogoMiMo}{MiMo-V2.6-Pro} & \avgrankcell{68.14}{\bestscore{7.154}} & \bestscore{0.879} & 36,345 & \bestscore{0.108} & \bestscore{3.432} & \bestscore{0.870} & \bestscore{0.473} & \bestscore{0.558} & \bestscore{0.757} & \bestscore{0.624} & \bestscore{0.374} & \secondscore{0.783} & \secondscore{3.352} & \secondscore{3.265} & \secondscore{3.236} & \secondscore{3.303} \\
    \modelentry{\modelLogoKimi}{Kimi K3.1 Pro} & \avgrankcell{64.96}{\secondscore{7.769}} & \secondscore{0.850} & \secondscore{25,448} & \secondscore{0.103} & \secondscore{3.367} & \secondscore{0.841} & \secondscore{0.467} & \secondscore{0.554} & \secondscore{0.734} & \secondscore{0.615} & \secondscore{0.372} & \bestscore{0.793} & \bestscore{3.455} & \bestscore{3.392} & \bestscore{3.352} & \bestscore{3.389} \\
    \modelentry{\modelLogoDeepSeek}{DeepSeek-V4.1-Flash}\nobreak\hspace{2pt}\frameinput & \avgrankcell{51.02}{10.462} & 0.824 & 65,570 & 0.097 & 3.212 & 0.815 & 0.440 & 0.530 & 0.713 & 0.598 & 0.354 & 0.757 & 3.216 & 3.179 & 3.141 & 3.153 \\
    \modelentry{\modelLogoGLM}{GLM-5.3-Flash} & \avgrankcell{45.05}{11.615} & 0.784 & 87,794 & 0.091 & 3.067 & 0.775 & 0.424 & 0.508 & 0.684 & 0.569 & 0.345 & 0.734 & 3.121 & 3.086 & 3.058 & 3.075 \\
    \modelentry{\modelLogoKimi}{Kimi K3} & \avgrankcell{38.48}{12.885} & 0.761 & 38,403 & 0.086 & 2.960 & 0.753 & 0.404 & 0.488 & 0.639 & 0.506 & 0.321 & 0.661 & 2.752 & 2.657 & 2.634 & 2.648 \\
    \modelentry{\modelLogoQwen}{Qwen3.8-27B} & \avgrankcell{26.33}{15.231} & 0.706 & 47,197 & 0.079 & 2.677 & 0.699 & 0.377 & 0.452 & 0.604 & 0.506 & 0.303 & 0.625 & 2.597 & 2.496 & 2.493 & 2.499 \\
    \modelentry{\modelLogoMiniMax}{MiniMax M3.1 Flash} & \avgrankcell{17.18}{17.000} & 0.677 & 43,818 & 0.078 & 2.602 & 0.671 & 0.363 & 0.426 & 0.582 & 0.489 & 0.291 & 0.568 & 2.389 & 2.389 & 2.346 & 2.383 \\
    \modelentry{\modelLogoMiniMax}{MiniMax M3} & \avgrankcell{12.40}{17.923} & 0.686 & \bestscore{11,880} & 0.068 & 2.560 & 0.680 & 0.342 & 0.418 & 0.544 & 0.410 & 0.280 & 0.481 & 1.879 & 1.873 & 1.775 & 1.784 \\
    \bottomrule
  \end{tabular*}
  \endgroup
\end{table*}

Tab.~\ref{tab:main_overall} compares 18 models using average ranks across 15 metrics, excluding ES and TC. As a robustness check, metrics with Spearman correlations of at least 0.8 are merged, changing model ranks by at most one position, with a correlation of 0.997 between the original and revised rankings.

\bpstart{Overall Performance}
Claude Opus 5.5 leads 14 of 15 ranked metrics with the lowest average rank of 1.067. Claude Sonnet 5.5 leads TQ and ranks second overall at 1.933, followed by GPT-6 Astra and Gemini 4 Argon. Among open-weight models, MiMo-V2.6-Pro ranks first and leads nine of ten objective quality, alignment, and fidelity metrics, while Kimi K3.1 Pro leads SF and all five subjective metrics. Execution success ranges from 0.597 to 0.958 and token consumption from 8,619 to 90,005.

\bpstart{Setting-Specific Patterns}
Tabs.~\ref{tab:main_t2v}--\ref{tab:main_v2v} report results for each input setting.
On T2V, Claude Opus 5.5 leads IF and CC, Claude Sonnet 5.5 leads MQ and VE, and GPT-6.1 Sol leads TQ and VD.
On I2V, Claude Opus 5.5 leads both image-fidelity metrics and all subjective metrics, while Gemini 4 Argon leads L-TVA.
GPT-6 Astra and GPT-6 Luna achieve similar image fidelity but differ considerably in IF and MQ, suggesting that keyframe fidelity alone does not ensure strong instruction following or motion quality.
On V2V, Claude Opus 5.5 leads all clip-fidelity and subjective metrics, while Claude Sonnet 5.5 achieves comparable clip fidelity but lower IF and CC, highlighting the gap between reference reconstruction and complete-video requirements.
Among open-weight models, DeepSeek-V4.1-Flash leads T2V and MiMo-V2.6-Pro leads I2V and V2V.

\section{Insights and Discussions}
\label{sec:insights}

We analyze direct video generation, model reliability, implementation choices and failures, task complexity, generation efficiency, V2V input sensitivity, and judge validity.

\subsection{Comparison with Direct Video Generation}
\label{sec:direct_generation}

We compare 18 MLLMs with Seedance 2.5~\cite{seedance25} on 60 tasks from 20 task groups, evenly split across T2V, I2V, and V2V, with target durations capped at 30 seconds by Seedance's generation constraint.
Seedance outperforms Claude Opus 5.5 on TQ, AQ, and L-TVA.
However, it scores 0.795 on IF and 3.217 on holistic video quality, exceeding the 18-MLLM averages of 0.754 and 3.149 but trailing Claude Opus 5.5's 0.946 and 4.008 in Tab.~\ref{tab:direct_generation}.
These results highlight code-based generation's potential for precise task fulfillment and subjective evaluation's diagnostic value for direct generation.

\begin{table}[t]
  \centering
  \captionsetup{skip=4pt}
  \caption{Subjective scores of selected models on 60 tasks. The average weights all 18 MLLMs equally.}
  \label{tab:direct_generation}
  \begingroup
  \footnotesize
  \setlength{\tabcolsep}{2pt}
  \renewcommand{\arraystretch}{1.0}
  \begin{tabular*}{0.95\columnwidth}{@{\extracolsep{\fill}}lcc@{}}
    \toprule
    \textbf{Model} & \textbf{Instruction Following} & \textbf{Holistic Video Quality} \\
    \midrule
    Claude Opus 5.5 & \textbf{0.946} & \textbf{4.008} \\
    GPT-6 Astra & 0.842 & 3.667 \\
    Kimi K3 & 0.845 & 3.571 \\
    \textbf{Seedance 2.5} & 0.795 & 3.217 \\
    \midrule
    18-MLLM average & 0.754 & 3.149 \\
    \bottomrule
  \end{tabular*}
  \endgroup
  \vspace{-1pt}
\end{table}

\subsection{Quality and Reliability Analysis}
\label{sec:conditional_quality}

\begin{figure}[!tbp]
  \centering
  \captionsetup{skip=4pt}
  \includegraphics[width=\columnwidth]{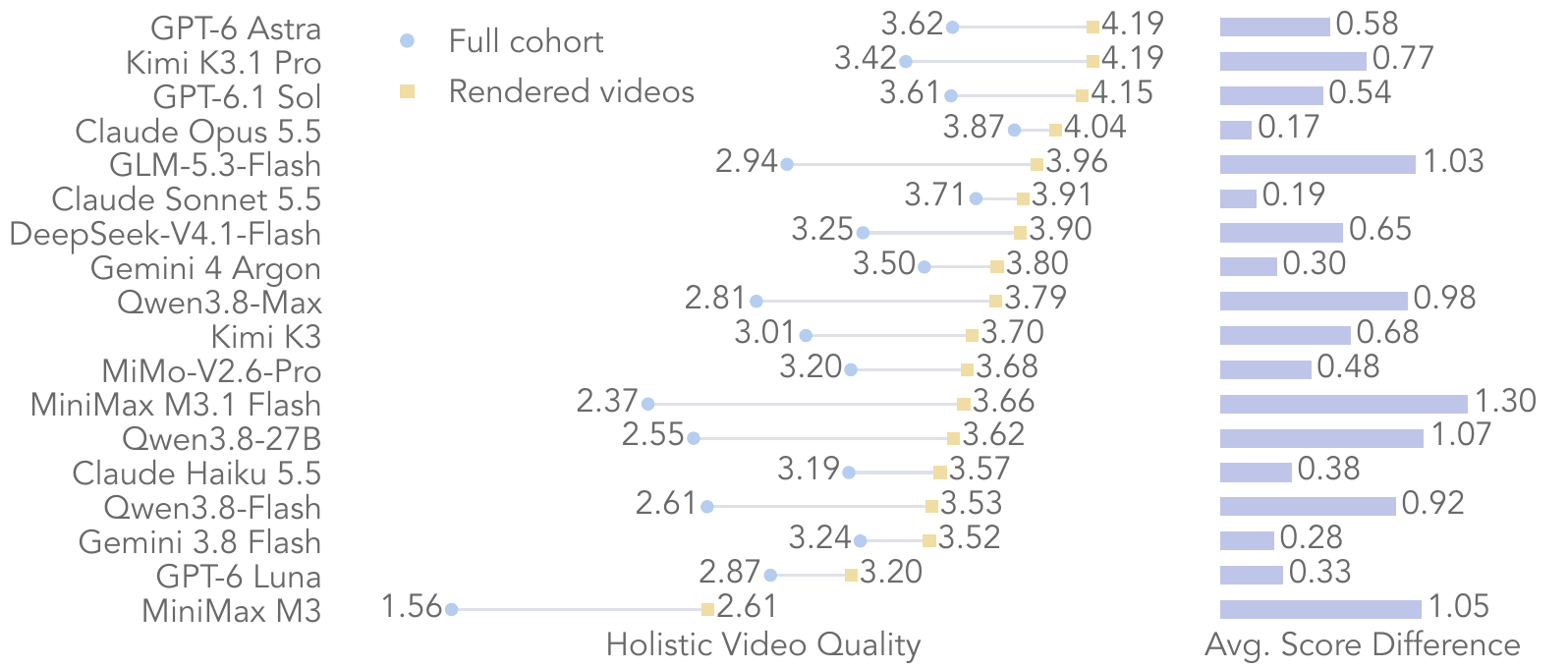}
  \caption{Mean holistic video quality over all tasks and successfully rendered videos. Bars show the gap.}
  \label{fig:insights_reliability}
  \vspace{-5pt}
\end{figure}

Given varying execution success rates in Sec.~\ref{sec:main_results}, Fig.~\ref{fig:insights_reliability} compares holistic video quality averaged across four metrics over all tasks and successful runs. GPT-6 Astra, Kimi K3.1 Pro, and GPT-6.1 Sol score 4.19, 4.19, and 4.15 on successful runs, exceeding Claude Opus 5.5 at 4.04, but failures lower their full-set means to 3.62, 3.42, and 3.61, leaving Claude Opus 5.5 in the lead at 3.87. This reversal motivates joint evaluation of quality and reliability. Meanwhile, successful code-based outputs score 0.987--0.993 on MS, indicating high local temporal predictability.

\subsection{Implementation and Failure Analysis}
\label{sec:failure_modes}
\begin{figure}[!tbp]
  \centering
  \captionsetup{skip=4pt}
  \includegraphics[width=\columnwidth]{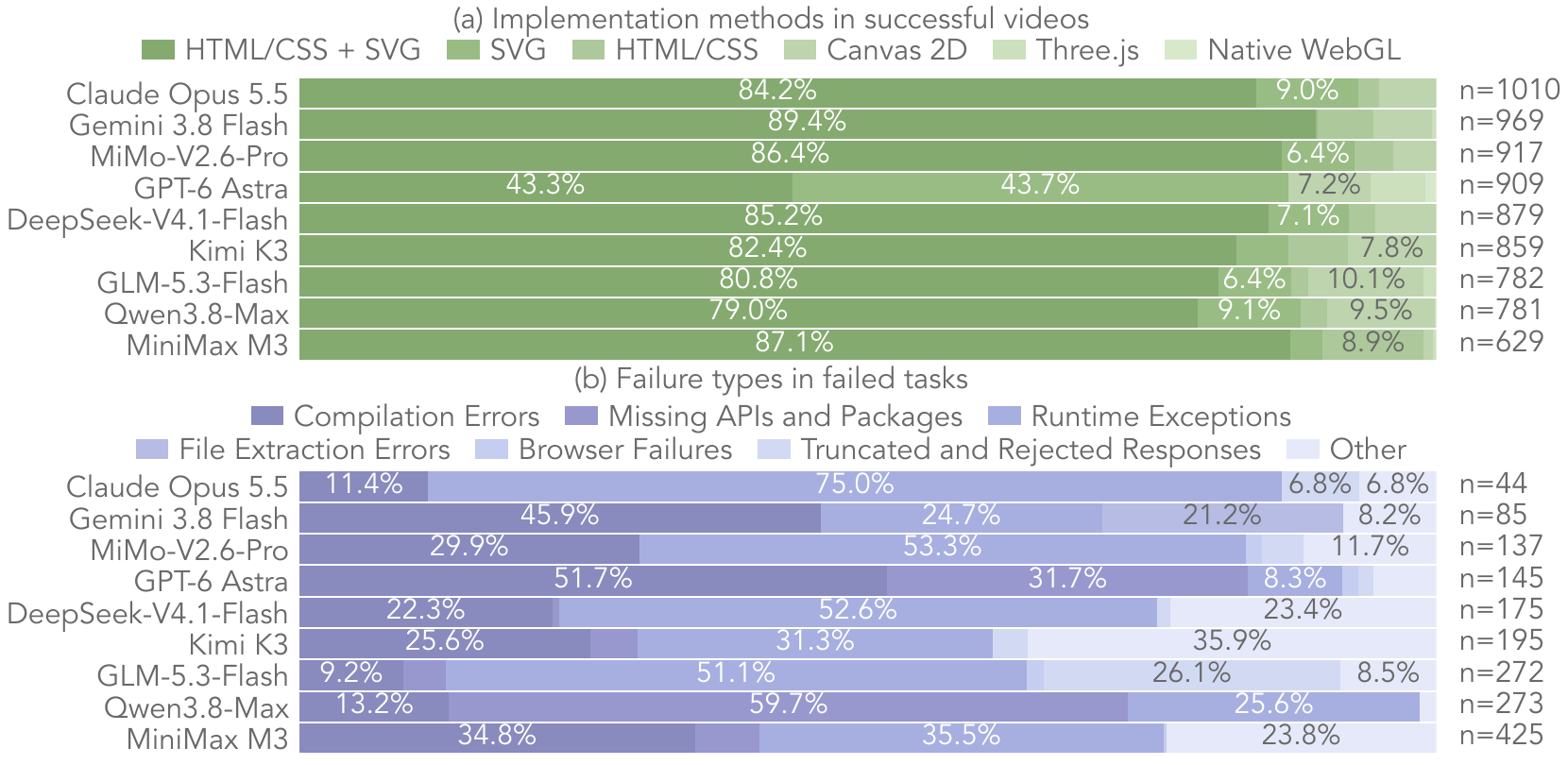}
  \caption{Implementation methods of successful videos (a) and failure types (b) for nine representative models.}
  \label{fig:insights_failures}
  \vspace{-3pt}
\end{figure}

We analyze implementation choices and failures across representative models in Fig.~\ref{fig:insights_failures}. HTML/CSS with SVG accounts for 79.0--89.4\% of successful outputs in eight of nine models, whereas GPT-6 Astra uses HTML/CSS with SVG and SVG in 43.3\% and 43.7\%, respectively. Compilation errors and missing APIs or packages account for 83.4\% of GPT-6 Astra's failures, compared with 59.7\% attributed to missing APIs or packages in Qwen3.8-Max and 75.0\% to runtime exceptions in Claude Opus 5.5.

\subsection{Complexity Analysis}
\label{sec:complexity_results}

\begin{figure}[!tbp]
  \centering
  \captionsetup{skip=4pt}
  \includegraphics[width=\columnwidth]{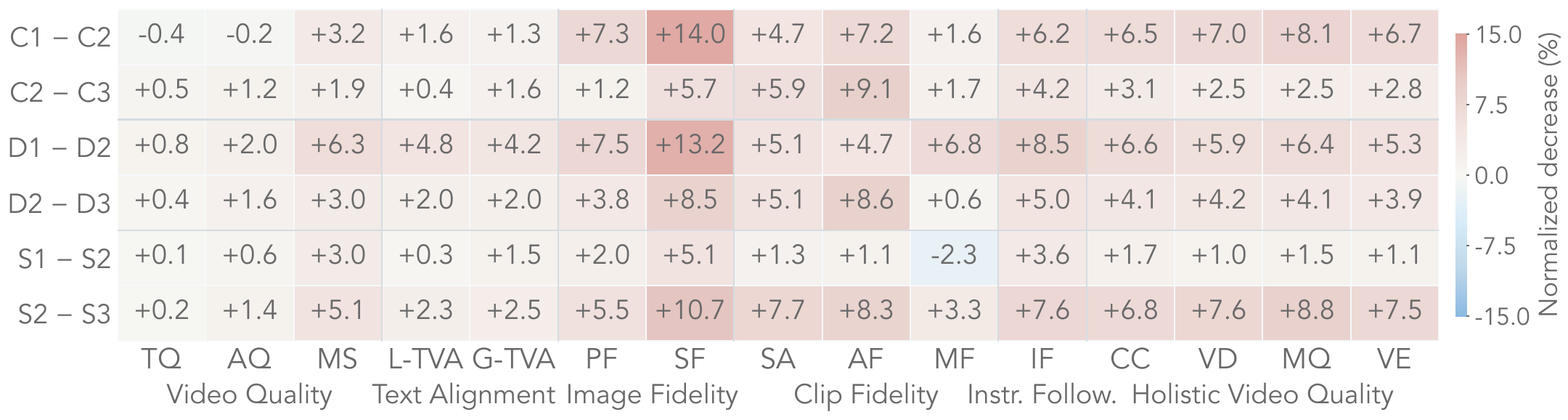}\caption{Performance changes between adjacent complexity levels, using the notation in Fig.~\ref{fig:dataset_complexity}. Values are normalized by metric scale.}
  \label{fig:insights_complexity}
  \vspace{-5pt}
\end{figure}

To examine how MLLMs handle increasing task complexity, we analyze construction, dynamic animation, and state dependency in Fig.~\ref{fig:insights_complexity}. Performance generally declines with increasing complexity, with SF dropping by 14.02\% and 13.16\% of its score range from L1 to L2 under construction and dynamic animation, respectively. Both dimensions show steeper declines from L1 to L2 than from L2 to L3 across most metrics, suggesting challenges even at moderate complexity. In contrast, state dependency shows larger declines from L2 to L3 across all 15 metrics, highlighting the difficulty of long-range dependencies.

\subsection{Generation Effort and Cost Analysis}
\label{sec:effort_cost}

\begin{figure}[!tbp]
  \centering
  \captionsetup{skip=4pt}
  \includegraphics[width=\columnwidth]{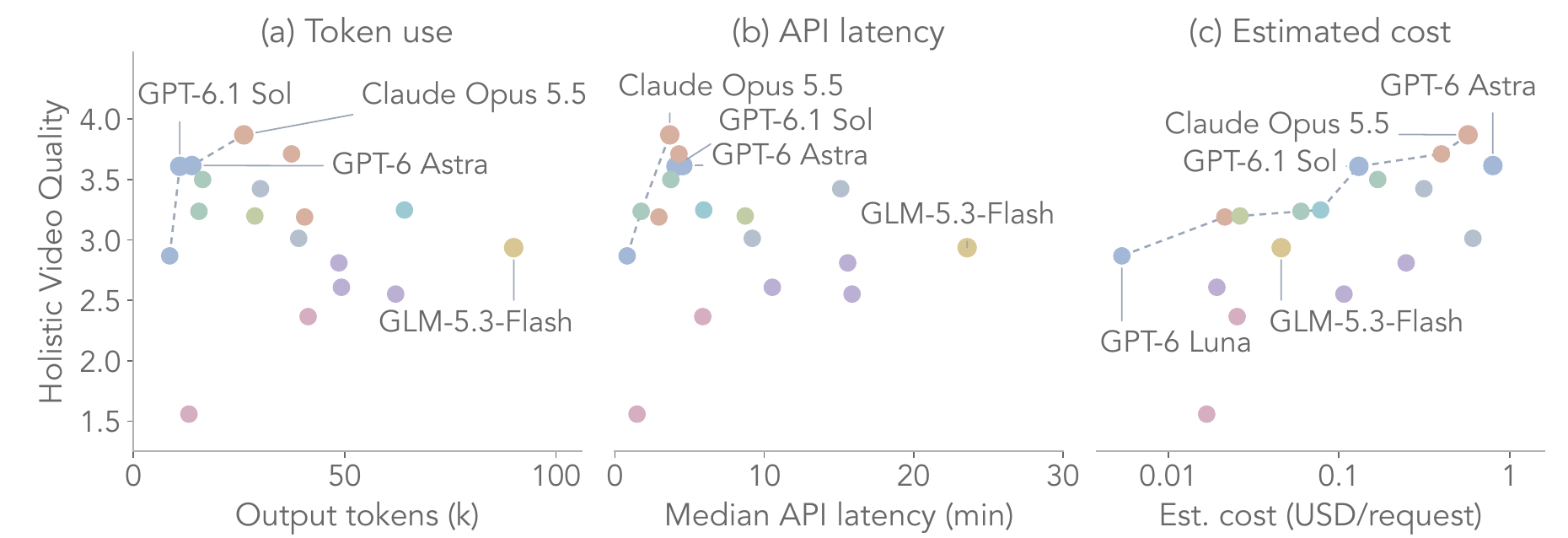}
  \caption{Output tokens (a), median API latency (b), and estimated API cost (c) versus mean holistic video quality across 18 models. Dashed lines mark Pareto frontiers.}
  \label{fig:insights_efficiency}
  \vspace{-3pt}
\end{figure}

We analyze holistic video quality against output token use, median API latency, and estimated API cost in Fig.~\ref{fig:insights_efficiency}. GLM-5.3-Flash consumes 90.0k output tokens per request with a median latency of 23.6 minutes, versus 13.9k and 4.6 minutes for GPT-6 Astra, yet scores lower at 2.94 versus 3.62. GPT-6.1 Sol nearly matches GPT-6 Astra at 3.61 versus 3.62 for approximately one-sixth of its API cost. Claude Opus 5.5 leads at 3.87 while costing less than GPT-6 Astra, showing that greater token consumption, latency, and spending do not guarantee higher video quality.

\subsection{Sensitivity Analysis of V2V Input Format}
\label{sec:v2v_input_format}

\begin{figure}[!tbp]
  \centering
  \captionsetup{skip=4pt}
  \includegraphics[width=\columnwidth]{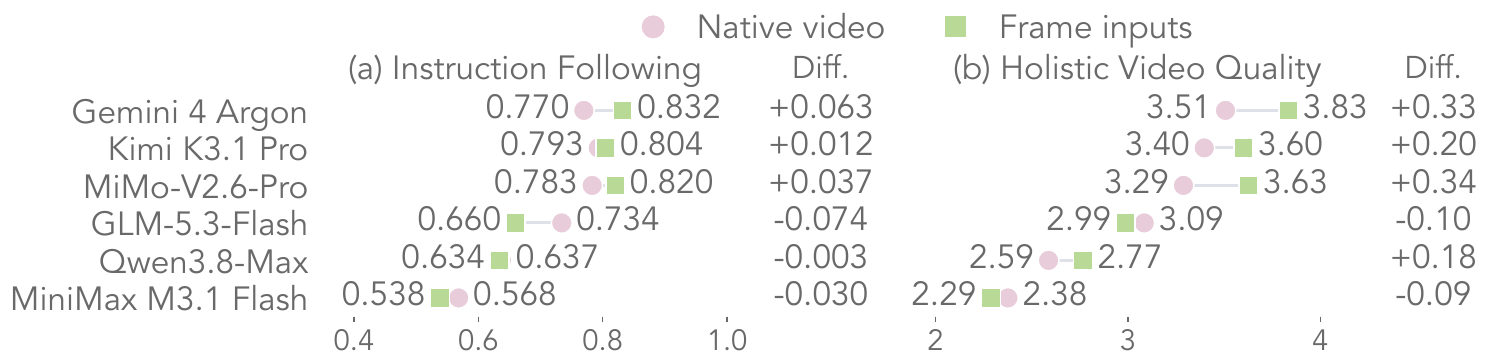}
  \caption{V2V input format sensitivity. (a) Instruction following. (b) Holistic video quality. Diff. indicates score differences.}
  \label{fig:v2v_input_format}
  \vspace{-5pt}
\end{figure}

We compare native video and 4 FPS frame inputs on identical V2V tasks for the top model from each of six families in Fig.~\ref{fig:v2v_input_format}. IF increases for three models and decreases for three, leaving the mean nearly unchanged from 0.714 to 0.715. Despite native video support, frame inputs improve holistic video quality for four models, including significant gains for Gemini 4 Argon and MiMo-V2.6-Pro, but reduce it for two, with the overall mean rising from 3.040 to 3.184. Replacing V2V scores with frame results leaves IF and holistic video quality rankings largely stable, with Spearman correlations of 0.977 for IF and 0.992 for holistic video quality.

\subsection{MLLM Judge Validation}
\label{sec:human_validation}

We validate Gemini 3.8 Flash on 300 outputs from six models across 50 shared tasks balanced by category and complexity, with three scoring runs each.
Under a protocol approved by our institution's internal ethics review, three independent raters assess IF and holistic video quality on 200 outputs, yielding ICC(A,3) values of 0.895 and 0.898, while Tab.~\ref{tab:judge_validation} shows strong human alignment and repeat stability.
To assess cross-judge consistency, we also use Qwen3.8-Max and MiMo-V2.6-Pro to score the same 200 outputs. Across the three judges, Cronbach's $\alpha$ is 0.943 for IF and 0.897 for holistic quality, with pairwise Pearson's $r$ ranging from 0.812 to 0.893 and 0.734 to 0.815, respectively.

\begin{table}[t]
  \centering
  \captionsetup{skip=4pt}
  \caption{Gemini 3.8 Flash judge validation. Human alignment uses Pearson's $r$, and repeat stability uses ICC(A,3) on 300 outputs.}
  \label{tab:judge_validation}
  \begingroup
  \footnotesize
  \setlength{\tabcolsep}{2pt}
  \renewcommand{\arraystretch}{1.12}
  \begin{tabular*}{0.9\columnwidth}{@{\extracolsep{\fill}}lcc@{}}
    \toprule
    \textbf{Metric} & \textbf{Human Alignment} & \textbf{Repeat Stability} \\
    \midrule
    Instruction Following & 0.796 & 0.983 \\
    Holistic Video Quality & 0.783 & 0.970 \\
    \bottomrule
  \end{tabular*}
  \vspace{-10pt}
  \endgroup
\end{table}

\FloatBarrier

\section{Conclusion}
\label{sec:conclusion}

We introduce \benchmark, a general-purpose benchmark for complete code-based video generation across T2V, I2V, and V2V, comprising 2,104 tasks and a dual-level evaluation framework. Evaluating 18 MLLMs reveals persistent challenges in execution reliability, instruction following, reference fidelity, and increasing task complexity.
We believe \benchmark offers insights into MLLM limitations and provides a testbed for advancing reliable, coherent, and high-quality code-based video generation.

\end{document}